\documentclass[a4paper]{cas-sc}

\usepackage[numbers]{natbib}
\usepackage[ruled,linesnumbered]{algorithm2e}
\usepackage{subfigure}
\usepackage{tabularx}
\usepackage{xurl}
\def\tsc#1{\csdef{#1}{\textsc{\lowercase{#1}}\xspace}}
\tsc{WGM}
\tsc{QE}
\begin{document}
\let\WriteBookmarks\relax
\let\printorcid\relax
\def\floatpagepagefraction{1}
\def\textpagefraction{.001}

\shorttitle{Supervised Contrastive Learning based Dual-Mixer Model for Remaining Useful Life Prediction}    

\shortauthors{En Fu, Yanyan Hu, Kaixiang Peng and Yuxin Chu}  

\title [mode = title]{Supervised Contrastive Learning based Dual-Mixer Model for Remaining Useful Life Prediction}  

% \tnotemark[<tnote number>] 


% \author[<aff no>]{<author name>}[<options>]

\author[1]{En Fu}
\ead{fuen@xs.ustb.edu.cn}
\credit{Conceptualization, Methodology, Writing – original draft, Software}

\author[1,2]{Yanyan Hu}
\ead{huyanyan@ustb.edu.cn}
\cortext[1]{Corresponding author}
\cormark[1]
\credit{Methodology, Supervision, Writing – review \& editing, Funding acquisition}

\author[3]{Kaixiang Peng}
\ead{kaixiang@ustb.edu.cn}
\credit{Supervision, Validation}

\author[1]{Yuxin Chu}
\ead{chuyuxin@xs.ustb.edu.cn}
\credit{Writing – review}
\affiliation[1]{organization={School of Intelligence Science and Technology, University of Science and Techonology Beijing},
            city={Beijing},
%          citysep={}, % Uncomment if no comma needed between city and postcode
            postcode={100083}, 
            country={China}}
\affiliation[2]{organization={Key Laboratory of Intelligent Bionic Unmanned Systems, Ministry of Education, University of Science and Technology Beijing},
            city={Beijing},
            postcode={100083}, 
            country={China}}
\affiliation[3]{organization={School of Automation and Electrical Engineering, University of Science and Technology Beijing},
            city={Beijing},
            postcode={100083}, 
            country={China}}

% \author[<aff no>]{<author name>}[<options>]

% Footnote of the second author

% Email id of the second author
% \ead{}



% \affiliation[<aff no>]{organization={},
%          citysep={}, % Uncomment if no comma needed between city and postcode

% \cortext[1]{Corresponding author}



\begin{abstract}
	The problem of the Remaining Useful Life (RUL) prediction, aiming at providing an accurate estimate of the remaining time from the current predicting moment to the complete failure of the device, has gained significant attention from researchers in recent years. In this paper, to overcome the shortcomings of rigid combination for temporal and spatial features in most existing RUL prediction approaches, a spatial-temporal homogeneous feature extractor, named Dual-Mixer model, is firstly proposed. Flexible layer-wise progressive feature fusion is employed to ensure the homogeneity of spatial-temporal features and enhance the prediction accuracy. Secondly, the Feature Space Global Relationship Invariance (FSGRI) training method is introduced based on supervised contrastive learning. This method maintains the consistency of relationships among sample features with their degradation patterns during model training, simplifying the subsequently regression task in the output layer and improving the model’s performance in RUL prediction. Finally, the effectiveness of the proposed method is validated through comparisons with other latest research works on the C-MAPSS dataset. The Dual-Mixer model demonstrates superiority across most metrics, while the FSGRI training method shows an average improvement of 7.00\% and 2.41\% in RMSE and MAPE, respectively, for all baseline models. Our experiments and model code are publicly available at \url{https://github.com/fuen1590/PhmDeepLearningProjects}.
\end{abstract}


% \begin{highlights}
% \item 1
% \item 2
% \item 3
% \end{highlights}

% Keywords

\maketitle

% Main text
\section{Introduction}\label{section1}
The prognostication of Remaining Useful Life (RUL), an essential component within the Prognostic and Health Management (PHM) system\cite{r1Compare2017-pe}\cite{Wang2023-sm}\cite{Wang2022-da}, has increasingly captured the focus of researchers. The primary goal of RUL prediction is to anticipate the remaining operational time or cycle lifespan of equipment or other designated objects, thereby offering valuable maintenance insights to maintenance personnel\cite{r2Han2021-tg}\cite{Li2023-ck}. The prevalence of data-driven approaches in RUL prediction has surged, primarily due to their robust nonlinear representation and commendable generalization capabilities\cite{r3Li2022-ou}. Recently, deep learning-based data-driven methods have exhibited renewed vitality. Such methods can fully leverage existing equipment monitoring data, achieving robust predictive performance in complex operating conditions without the need to delve into the intricacies of equipment mechanisms. Numerous studies have already demonstrated the effectiveness of these methods when confronted with abundant equipment monitoring data\cite{r4Zhang2022-zn}\cite{r5Zhang2023-xb}\cite{r6Zhang2023-kc}\cite{r7Xu2023-cl}\cite{Wang2022-lj}.

Deep learning methods aim to uncover information at different levels hidden within data and encode these insights into a high-dimensional feature space, revealing characteristics that cannot be directly expressed in the original data dimensions, referred to as "features". Temporal and spatial dimensions are widely regarded as fundamental dimensions for characterizing equipment monitoring data\cite{r6Zhang2023-kc}\cite{r7Xu2023-cl}\cite{r8Sateesh_Babu2016-cs}. Given the inherent temporal characteristics of equipment monitoring data, the temporal feature mining module is considered a foundational element and is widely employed as a basic module in numerous methods. Spatial features are equally crucial in certain scenarios, especially when there are multiple monitoring points or the equipment is complex\cite{r9Ren2021-yv}\cite{Li2023-hb}. In such cases, the spatial relationships among data variables become more intricate, and a single variable alone cannot accurately describe the characteristics of the equipment. Therefore, spatial features are utilized to compensate for the limitations of temporal features, collectively serving as descriptors for the equipment. However, we have identified two primary problems in existing methods:

1. RUL prediction typically involves multivariate equipment monitoring time series data, such as multiple sets of vibration signals in different directions, etc. To fully utilize the information provided by temporal and spatial dimensional, multi-dimensional feature fusion is a fundamental module in data-driven methods. However, most current methods are constrained by the fact that temporal and spatial features are extracted from different modules. The differences in feature structures result in feature fusion being predominantly accomplished through simple operations, such as addition or concatenation, thus lacking flexibility. This makes it challenging for the model to distinguish the importance of different features. Therefore, investigating flexible feature fusion is crucial for enhancing the flexibility and robustness of data-driven methods.

\begin{figure}[ht]
      \centering
 		\includegraphics[scale=0.5]{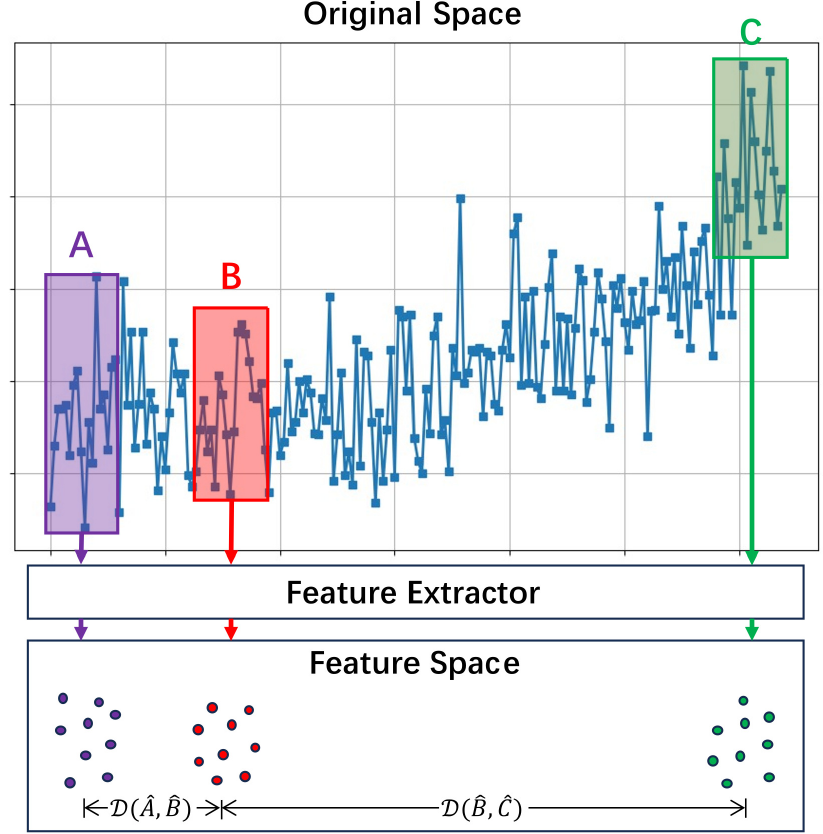}
 	  \caption{Illustration of progression relationship between samples for equipment degradation process.}\label{fig1}
\end{figure}

2. Currently, the RUL prediction problems are often transformed into regression problems, regressing the corresponding RUL values using samples from a single time window. Common approaches individually regress one RUL value with one sample for training, without taking into account the potential relationships between samples. This relationship becomes more apparent when the degradation intervals are sufficiently large. For example, as illustrated in Figure \ref{fig1}, samples A, B, and C are constructed using a sliding window. Sample B is mapped into the feature space and subsequently regressed to its corresponding RUL. Obviously, the degradation time between samples A and B is much shorter than that between B and C. In the high-dimensional feature space, the similar relationship should be remained to simplify subsequent regression task in the output layer. The excessive freedom in the feature extractor of current models hinders the stability of the final output results. Therefore, correctly incorporating the global relationship between samples as a constraint in the feature extractor optimization process will enhance the performance of existing data-driven methods in RUL prediction.

Addressing the two main issues highlighted above, this study has undertaken the following initiatives:

1. To resolve issue 1, and drawing inspiration from the recent success of the MLP-Mixer architecture in time series prediction, a lightweight Dual-path Mixer, referred to as the Dual-Mixer, has been devised in this work. The model utilizes homogeneous modules to extract spatial and temporal features at the same time. This not only decouples the process of extracting spatial and temporal features but also maintains structural similarity among these features, facilitating ease of fusion. Additionally, to enable the model to fully utilize features at different scales for optimal performance, a lighter gate mechanism, compared to attention mechanism, is designed in the deep structure of the network. This achieves a dynamic layer-wise feature fusion process.

2. This paper proposes a model training method called Feature Space Global Relationship Invariance (FSGRI) based on supervised contrastive learning to solve issue 2. This method utilizes the relative degradation relationships among samples within the original data space as constraints on the distribution in the feature space. It significantly enhances output stability and simplifies the regression task in the output layer. Positive and negative sample pairs are constructed within the same degradation unit. It specifies optimization weights for different negative samples through negative sample RUL values, providing global degradation position information during the model optimization process. Additionally, it utilizes noise augmentation for enhanced sample diversity, thus improving model robustness. This method is not restricted to specific model structures, showcasing high versatility. Experimental validations on a multitude of existing methods corroborate the effectiveness of the global information training method.

In addition, comparative assessments against the latest RUL prediction methods on a popular public dataset, C-MAPSS dataset, validate the superiority of the proposed methods in this paper.

\section{Realted Work}\label{section2}
\textbf{RUL Prediction:} In recent years, with the improvement of data collection capabilities, advancements in sensor technology, and the widespread adoption of big data techniques, data-driven RUL prediction methods have gradually become the mainstream of research. Among them, deep learning-based approaches can fully exploit internal information in the data, offering strong robustness, generality, and ease of use. The latest achievements primarily revolve around enhancing the model's ability for temporal feature extraction, spatial-temporal fusion feature extraction, and improvements in attention mechanisms. For instance, Zhang et al.\cite{r4Zhang2022-zn} proposed a temporal attention mechanism combined with bidirectional Gated Recurrent Units (GRUs) for temporal feature extraction, aiming to construct a more accurate RUL prediction model. Zhang and Li et al.\cite{r5Zhang2023-xb} based their work on self-attention modules, incorporating two types of sparse self-attention mechanisms, achieving state-of-the-art results on the C-MAPSS dataset. Zhang and Tian et al. \cite{r6Zhang2023-kc} integrated one-dimensional CNN with bidirectional GRU networks to simultaneously extract spatiotemporal features, significantly improving the model's accuracy in RUL prediction. Pei et al.\cite{Pei2023-ev} introduced an interactive prediction framework that utilizes stacked autoencoders for constructing health indicators and incorporates a nonlinear degradation model. The effectiveness and superiority of the method were demonstrated through two case studies involving turbofan engines. Ren et al.\cite{r9Ren2021-yv} comprehensively employed autoencoder structures, CNN, and LSTM to achieve adaptive spatiotemporal feature extraction for battery data, used for predicting battery RUL. While these works attempt to enhance the accuracy and applicability of deep learning methods in RUL prediction from different perspectives, we observe that the RUL prediction problem can be extended to a regression problem for multivariate time series. Therefore, starting from the latest time series analysis models can further improve the performance of data-driven methods in RUL prediction.

\textbf{Multivariate Time Series Analysis:} Recently, research in time series analysis has predominantly focused on Transformer architectures, attention mechanisms, and Multi-Layer Perceptron (MLP) models, which have significantly influenced the development of RUL prediction in various domains\cite{r5Zhang2023-xb}\cite{r9Ren2021-yv}\cite{r10Li2022-zp}\cite{r11Zhang2024-aw}. While Transformers exhibit excellent sequential modeling capabilities, their bulky architecture and parameter scale often pose challenges in industries due to difficulties in lightweight implementation\cite{r5Zhang2023-xb}. Therefore, there is a growing interest in more lightweight MLP-like models for time series analysis. Zhang et al.\cite{r12Zhang2022-cu} decomposed time series into trend and cyclical components, utilizing a lightweight MLP network for prediction. Zeng et al.\cite{r13Zeng2023-ft} discovered that simple MLP models can achieve, and sometimes surpass, the performance of complex Transformer networks in handling time series data, paving the way for new avenues in time series analysis research. TS-Mixer\cite{r14Ekambaram2023-bk} adopted the approach of MLP-Mixer\cite{r15Tolstikhin2021-xm} from computer vision, employing MLPs for feature extraction across Patch, Spatial, and Temporal dimensions. It achieved outstanding results in time series prediction and featured a simpler structure with fewer parameters compared to Transformer architectures. Similarly, Google proposed Tsmixer\cite{r16Chen2023-ut}, a pure MLP-based architecture for multivariate time series prediction, highlighting the robust potential of pure MLP structures in processing multivariate time series data. Therefore, this paper will explore the application of pure MLP structures in data-driven RUL prediction.

\section{Methodology}\label{section3}
This chapter will first introduce the basic definition of the RUL prediction problem in Section \ref{section31}. Subsequently, Section \ref{section32} will provide detailed insights into the proposed Dual-path Mixer model. Finally, Section \ref{section33} will present the training method based on the FSGRI constraint.
\subsection{Notations}\label{section31}
The problem description for the RUL prediction problem discussed in this paper is as follows. $\mathcal{X}^{l \times m}_i$ represents the $i$-th multivariate equipment monitoring time series sample with a length of $l$ and $m$ variables.$\mathcal{F}(\cdot)$ denotes the feature extractor, $\mathcal{A}(\cdot)$ represents the output layer used for regression, $\mathcal{Y}_i$ denotes the true RUL value for the $i$-th sample, and $\hat{\mathcal{Y}}_i$ represents the predicted RUL value for the $i$-th sample. The expression for the process RUL prediction can be formulated as follows:

\begin{equation} \label{eq1}
	\hat{\mathcal{Y}}_i = \mathcal{A}(\mathcal{F}(\mathcal{X}^{l \times d}_i))
\end{equation}

\subsection{Dual-path Mixer for Remaining Useful Life Prediction} \label{section32}
\subsubsection{Basic Block} \label{section_basic_block}
\begin{figure}[ht]
	\centering
	\includegraphics[scale=0.5]{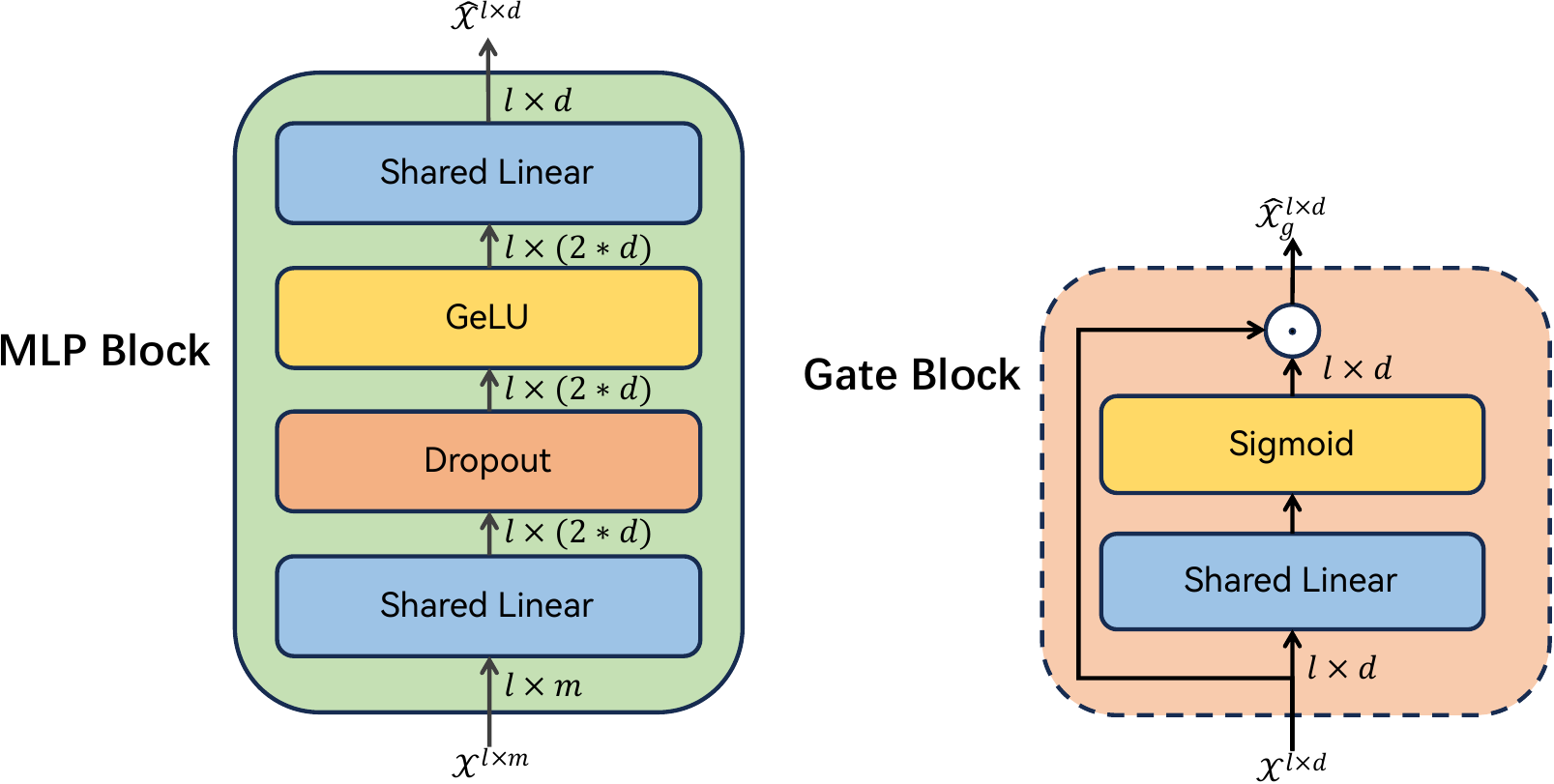}
	\caption{Basic MLP Block and Gate Block in the proposed Dual-Mixer model.}\label{fig2}
\end{figure}

In recent years, MLP Mixer has demonstrated powerful processing capabilities in time series feature extraction\cite{r14Ekambaram2023-bk}\cite{r16Chen2023-ut}. Inspired by models like MLP Mixer, which are based on a pure MLP architecture, the construction of a spatial-temporal homogeneous feature extractor has become feasible. This paper utilizes a standard MLP module as the fundamental unit for feature extraction, as illustrated on the left side of Figure \ref{fig2}. The matrix shapes of the output features for each layer are indicated in the figure. The core of this module is a linear transformation layer with shared parameters across channels, represented mathematically as follows:
\begin{equation}
	{\hat{\mathcal{X}}}^{l \times d} = GeLU( \mathcal{X}^{l \times m}*W_{m1})*W_{m2}
\end{equation}
where $W_{m1} \in \mathbb{R}^{m \times {({2*d})}}$ and $W_{m2} \in \mathbb{R}^{{({2*d})} \times d}$ are the parameter matrices to be optimized, $d$ is the feature dimension of the model, $GeLU$ stands for Gaussian Error Linear Units\cite{r17Hendrycks2016-dc} activation function, and $(*)$ denotes matrix multiplication. Additionally, a simple lightweight gating unit is designed for subsequent dynamic feature fusion, as illustrated on the right side of Figure \ref{fig2}. Inspired by the LSTM gating mechanism\cite{r18Hochreiter1997-ok}, this module eliminates the need for a complex attention mechanism, allowing the removal of interference components from features and providing the model with feature selection capabilities. Its mathematical form is as follows:
\begin{equation}
	\hat{\mathcal{X}}^{l \times d}_g = Sigmoid(\mathcal{X}^{l \times d} * W_g) \odot \mathcal{X}^{l \times d}
\end{equation}
where $\odot$ represents the element-wise product (Hadamard product), and $W_{g} \in \mathbb{R}^{d \times d}$ is the parameter matrix to be optimized. Essentially, this module learns a weight matrix with the same shape as the input features and filters the original input features with the weight matrix through Hadamard product.

\subsubsection{Dual-path Mixer Layer} \label{section_MLP_layer}
\begin{figure}[ht]
	\centering
	\includegraphics[scale=0.5]{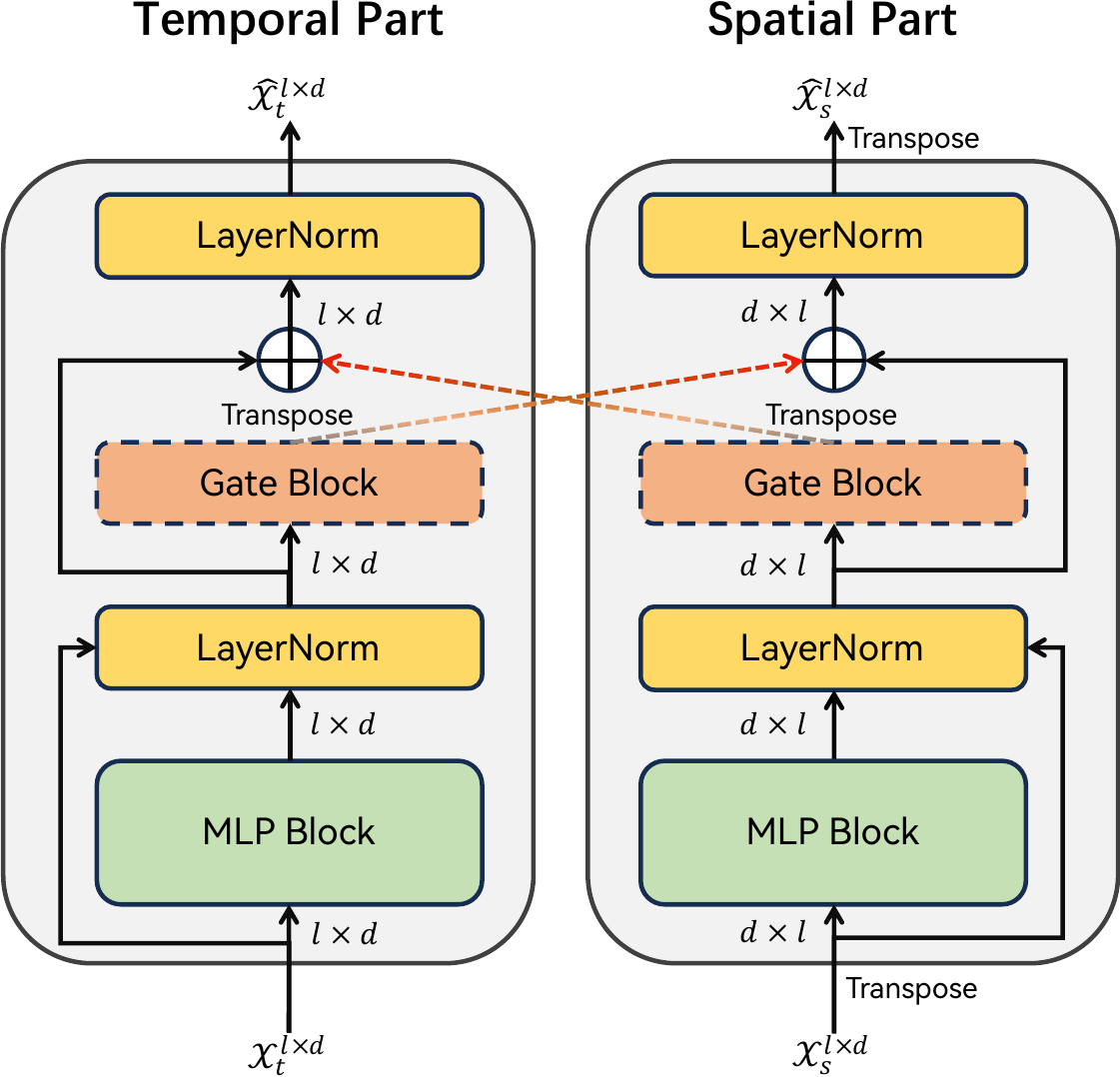}
	\caption{Proposed Dual-Mixer Layer structure.}\label{fig3}
\end{figure}

Through the basic modules introduced in the previous section, the proposed Dual-path Mixer Layer (DML) can be constructed. The goals of designing DML are threefold: 1) Construct a homogeneous feature extractor to make the feature structures of different dimensions similar without introducing additional priors, maintaining the generality of the structure. 2) Achieve flexible feature fusion and interaction, endowing the final model with feature selection capability. 3) Be easily stackable into a deep architecture, maintaining the flexibility of DML and enabling the final model to achieve stronger non-linear mapping capabilities with a limited number of parameters. 
Based on the above three main goals, the DML structure is designed as illustrated in Figure \ref{fig3}.
Firstly, DML consists of two parts: the temporal part and the spatial part (but not limited to two parts; more parallel parts can be constructed when additional feature dimensions are available). The core difference between the two parts lies in the dimensionality of the processing of their respective input features. Due to the shared linear mapping layers across the first dimension in the MLP Block and Gate Block, as introduced in Section \ref{section_basic_block}, we can achieve distinct dimensions of focus for the two parts of DML by applying a transpose operation. By processing them separately on two sets of different dimensions, the model can capture diverse aspect features of the data. Moreover, since the module architecture used by the two parts of the feature extractor is entirely identical, differing only in parameter matrices, goal 1) is effectively accomplished.

Secondly, the interaction and fusion of features from the two parts occur at the red connection lines shown in the Figure \ref{fig3}, referred to as "interaction connections." Similar to the residual connection exchanging information in the depth direction of the network, interaction connections exchanging information in the width direction of the network. To control the flow of features between the two parts, a Gate Block is employed for feature filtering before feature fusion, achieving goal 2). 

Finally, it's worth noting that the features filtered by the Gate Block are used only for feature exchange and not for subsequent processing steps within its own part. This is done to avoid the gradient being propagated only through the Sigmoid function used by the Gate Block during backpropagation, mitigating the impact of the Sigmoid's gradient saturation region on the convergence speed of the model. Additionally, after the MLP Block and feature fusion, a combination of residual connection and LayerNorm is used to stabilize the gradient values during backpropagation\cite{r19Xu2019-vs}. This allows the model to be stacked into a deep and narrow structure, gaining stronger non-linear mapping capabilities without affecting the convergence speed, achieving goal 3).

In summary, the overall workflow of the DML module can be described as follows:
\begin{equation}
	\mathcal{Z}_{t}^{l \times d} = LN\left( M_{1}\left( \mathcal{X}_{t}^{l \times d} \right) + \mathcal{X}_{t}^{l \times d} \right)
\end{equation}
\begin{equation}
	\mathcal{Z}_{s}^{d \times l} = LN\left( M_{2}\left( \left( \mathcal{X}_{s}^{l \times d} \right)^{T} \right) + \mathcal{X}_{s}^{d \times l} \right)
\end{equation}
\begin{equation}
	{\hat{\mathcal{X}}}_{t}^{l \times d} = LN\left( G_{2}\left( \mathcal{Z}_{s}^{d \times l} \right)^{T} + \mathcal{Z}_{t}^{l \times d} \right)
\end{equation}
\begin{equation}
	{\hat{\mathcal{X}}}_{s}^{l \times d} = LN\left( {G_{1}\left( \mathcal{Z}_{t}^{l \times d} \right)^{T} + \mathcal{Z}_{s}^{d \times l}} \right)^{T}
\end{equation}
where $\mathcal{X}^{l \times d}$ and $\mathcal{X}^{l \times d}$ represent the input matrices for the temporal and spatial parts, respectively. $M_1$ and $M_2$ denote the MLP Blocks in the temporal and spatial parts, while $G_1$ and $G_2$ refer to the Gate Blocks in the temporal and spatial parts. $LN(\cdot)$ represents the LayerNorm module.

It is important to note that the feature dimensionality $d$ of the MLP Block $M_1$ and Gate Block $G_1$ in the Temporal Part is a hyperparameter specified manually. However, in the Spatial Part, the feature dimensionality of MLP Block $M_2$ and Gate Block $G_2$ is denoted as $l$, representing the time length of the input sample data, and it is not a hyperparameter but rather derived from the input sample data $\mathcal{X}^{l \times d}$.

\subsubsection{Dual-path Mixer Model} \label{section_dp_mixer_model}
\begin{figure}[ht]
	\centering
	\includegraphics[scale=0.5]{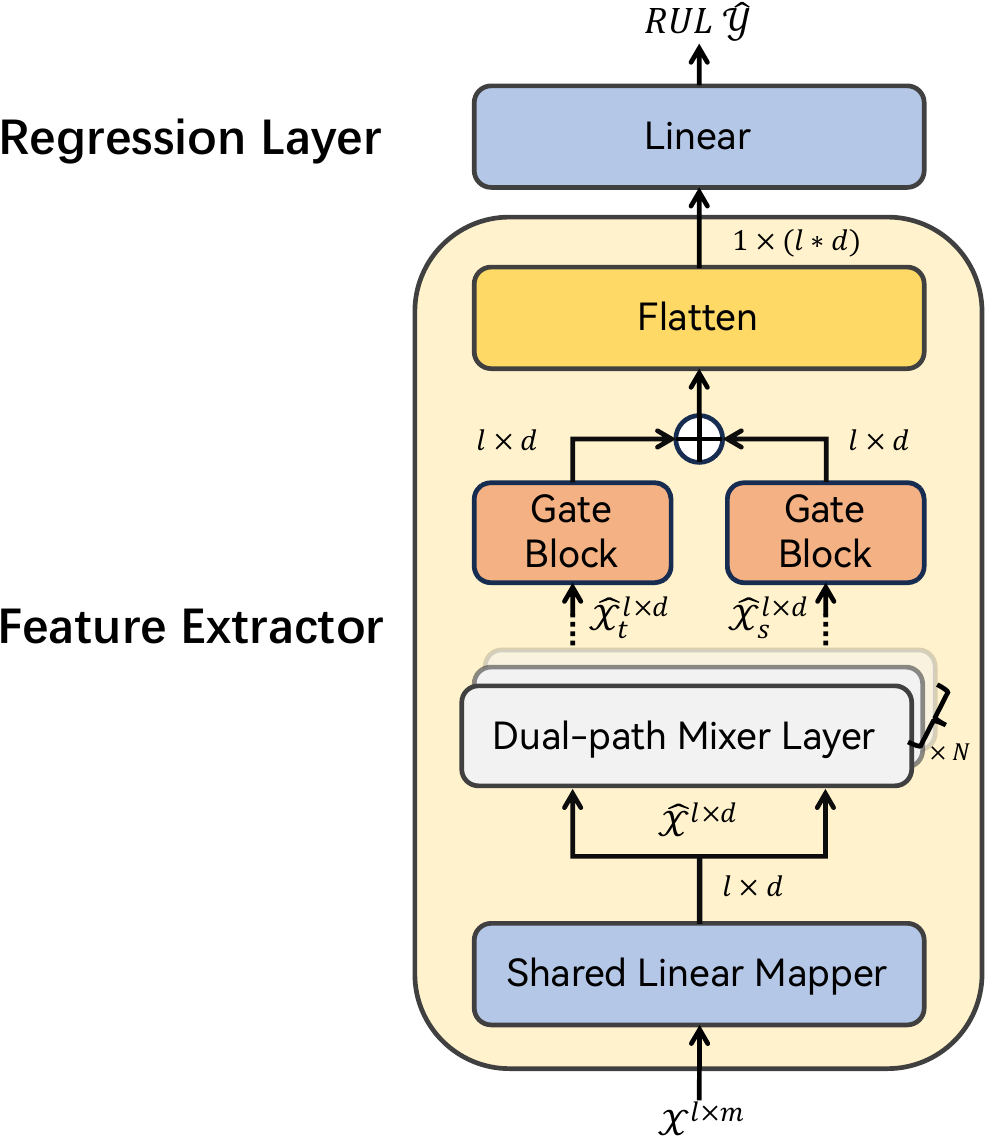}
	\caption{Dual-path Mixer Model structure.}\label{fig4}
\end{figure}
Through the basic modules and DML modules mentioned above, a RUL prediction model called Dual-path Mixer (Dual-Mixer) is constructed for equipment multivariate monitoring time series data, as illustrated in Figure \ref{fig4}. The model is divided into a feature extractor and a regression layer. The feature extractor is used to map input samples from the raw data space to a high-dimensional feature space, while the regression layer is employed to regress the extracted high-dimensional features to the RUL.
The feature extraction part consists of a linear mapping layer, multiple DML layers, and two independent Gate Blocks. The role of the linear mapping layer is to initially map the input samples to the model's feature dimensions, and its mathematical form is as follows:
\begin{equation}
	{\hat{\mathcal{X}}}^{l \times d} = \mathcal{X}^{l \times m}*W_{m}
\end{equation}
where $W_{m} \in \mathbb{R}^{m \times d}$ represents the parameters to be optimized and d is the feature dimension. Subsequently, for the first layer of DML, both the temporal and spatial parts receive input from the same input matrix $\hat{\mathcal{X}}^{l \times d}$, where the spatial part is distinguished from the temporal part through the transposition operation illustrated in Figure \ref{fig3}. The computational process for each layer of DML is described in Section \ref{section_MLP_layer}.
Finally, the last layer of the DML module outputs two distinct features: temporal features $\mathcal{X}^{l \times d}_t$ and spatial features $\mathcal{X}^{l \times d}_s$. Each feature undergoes further filtering through a Gate Block before being summed together to form the final feature. As the ultimate feature integrates both temporal and spatial characteristics, it can be conveniently flattened for subsequent processing by the regression layer.
The regression part is composed of a linear mapping layer, and its mathematical form is as follows:
\begin{equation}
	\hat{\mathcal{Y}} = \mathcal{X}^{1 \times (l*d)} * W_r 
\end{equation}
where $W_r \in \mathbb{R}^{(l*d) \times 1}$.

\subsection{Feature Space Global Relationship Invariance Training} \label{section33}
\subsubsection{Feature Space Global Relationship Invariance} \label{section_fsgri}
The proposed FSGRI aims to preserve the relative spatial relationships in the original data space within the feature space. The envisioned relationships between features established by FSGRI can be described as:
\begin{equation} \label{fsgri_eq}
	s(\mathcal{Z}_i, \mathcal{Z}_i)>s(\mathcal{Z}_i, \mathcal{Z}_{i+1})> \dots >s(\mathcal{Z}_i, \mathcal{Z}_t)
\end{equation}
where $\mathcal{Z}_i$ represents the high-dimensional features of the $i$-th sample, $s$ is a scoring function, and its output score $s_{i,k}$ quantifies the degree of matching between sample features. A higher score indicates a higher degree of match. In this paper, $s$ is the cosine similarity function, and $i \in [1, t]$, arranged in chronological order. This corresponds to the RUL of the samples, satisfying:
\begin{equation} \label{fsgri_y_eq}
	\mathcal{Y}_i>\mathcal{Y}_{i+1}>\dots >\mathcal{Y}_t
\end{equation}
This relationship indicates that samples with close distances in the original space have similar features. Conversely, samples that are farther apart exhibit greater feature differences, and these differences vary with distance. This constraint is employed to achieve continuous encoding in the feature space, simplifying the regression challenge for RUL.

Contrastive learning can constrain the encoding distances of features between different samples from the perspective of the feature space and is widely applied in various pre-training and classification tasks\cite{r20He2020-hc}. Due to its inherent ability to impose constraints on the feature space, contrastive learning is selected to implement FSGRI. The next section will first introduce the construction method for positive and negative sample pairs.

\subsubsection{Gaussian Threshold Sampling method} \label{section_guassian_sampling}
\begin{figure}[ht]
	\centering
	\includegraphics[scale=0.5]{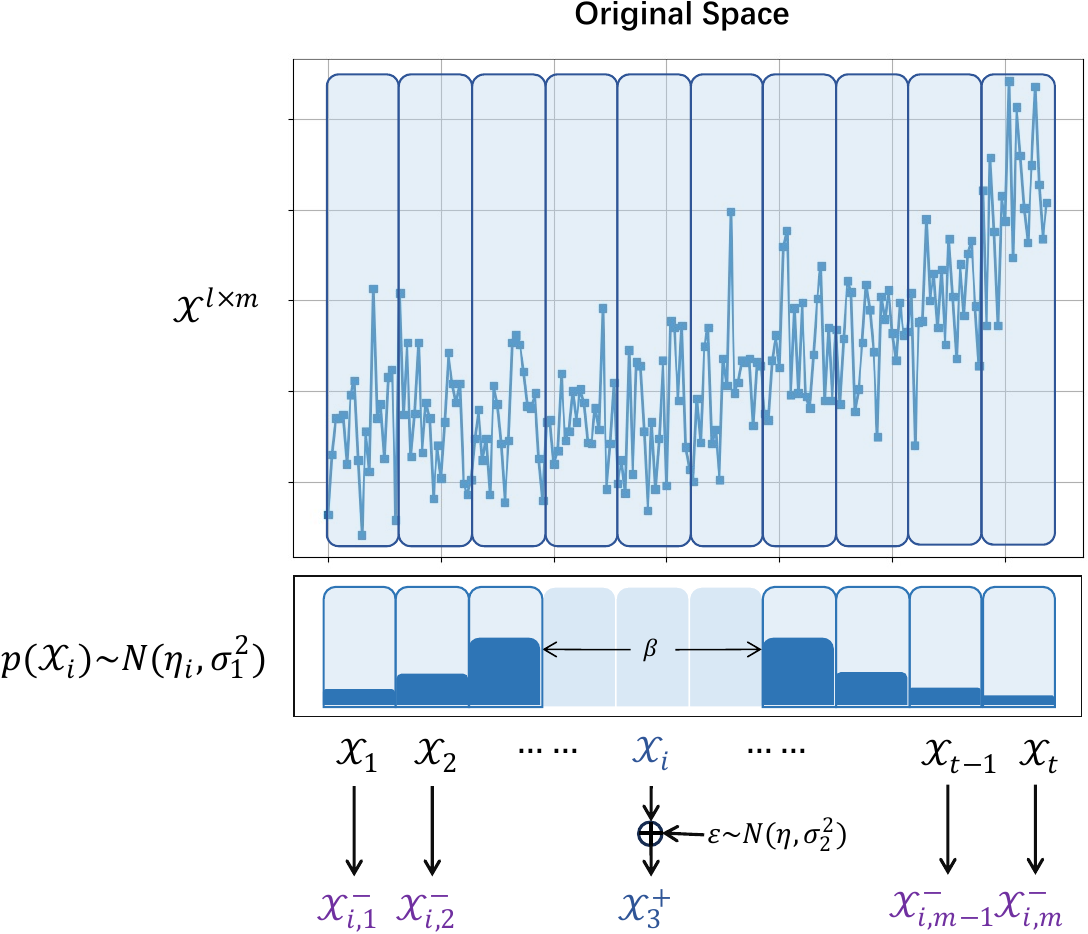}
	\caption{The illustration of proposed Gaussian Threshold sampling method.}\label{fig5}
\end{figure}

The strategy for constructing positive and negative sample pairs significantly impacts learning effectiveness. Inspired by the TNC\cite{r22Tonekaboni2021-sn} sampling method, we propose a Gaussian Threshold Sampling method for RUL prediction to construct positive and negative sample pairs, as illustrated in Figure \ref{fig5}.
Given the complete degradation data $\mathcal{X}^{l \times m}$ for one device, a sliding window of size $w$ and step size $sl$ is used to generate $t$ samples $[\mathcal{X}_1,\mathcal{X}_2,\dots,\mathcal{X}_t]$, where $\mathcal{X}_i \in \mathbb{R}^{w \times m}$ . During the training phase, when sample $\mathcal{X}_i$ is chosen as the training sample (anchor sample), the probability $p(\mathcal{X}_k)$ of sampling the other samples as negative samples is as follows:
\begin{align} \label{gaussian_prob_eq}
	\begin{cases}
		p(\mathcal{X}_k)\sim N(\eta_i, \sigma_{1}^{2}),&k<(i-\frac{t\beta}{2})~ or~ k>(i+\frac{t\beta}{2})\\
		p(\mathcal{X}_k)=0,&(i-\frac{t\beta}{2}) \leq k \leq (i+\frac{t\beta}{2})
	\end{cases}
\end{align}
where $\beta$ is the threshold coefficient, indicating the non-sampling range centered around $i$. The sampling probability of other samples follows a Gaussian distribution $N(\eta_i,\sigma_{1}^{2} )$, with the distribution always centered at the sampling index $i$. In other words, samples around $\mathcal{X}_i$ are more likely to be sampled as negative samples, but the threshold coefficient $\beta$ ensures that the sampling distance is not too close. Finally, by non-repetitively sampling, $m$ negative samples $[\mathcal{X}_{i,1}^{-}, \dots, \mathcal{X}_{i,m}^{-}]$ can be obtained.

The construction of positive sample is achieved by adding Gaussian noise to the anchor sample $\mathcal{X}_i$:
\begin{equation} \label{positive_sample_eq}
	\mathcal{X}_{i}^{+} = \mathcal{X}_i + \epsilon,~~\epsilon \sim N(0, \sigma_{2}^{2})
\end{equation}

\subsubsection{Distance Weighted InfoNCE} \label{section_DWI}
InfoNCE\cite{r23Van_den_Oord2018-ua} is a common contrastive learning loss function that leverages positive and negative sample pairs to establish relationships between features:
\begin{align}
	\begin{split}
		\mathcal{L}_{InfoNCE} &= - \log\left( \frac{\exp\left( \frac{s\left( {\mathcal{Z}_{i},\mathcal{Z}_{i}^{+}} \right)}{\tau} \right)}{{\sum_{k! = i}{\exp\left( \frac{s\left( {\mathcal{Z}_{i},\mathcal{Z}_{i,k}^{-}} \right)}{\tau} \right)}} + {\exp\left( \frac{s\left( {\mathcal{Z}_{i},\mathcal{Z}_{i}^{+}} \right)}{\tau} \right)}} \right)\\
		&= - \log\left( \frac{\exp\left( \frac{s_{i,i}}{\tau} \right)}{{\sum_{k \neq i}{\exp\left( \frac{s_{i,k}}{\tau} \right)}} + {\exp\left( \frac{s_{i,i}}{\tau} \right)}} \right)
	\end{split}
\end{align}
where $\mathcal{Z}_i$ and $\mathcal{Z}_{i}^{+}$ are the feature encodings in the feature space for the anchor sample $\mathcal{X}_i$ and the positive sample $\mathcal{X}_{i}^{+}$ respectively. $\mathcal{Z}_{i,k}^{-}$ represents the feature encoding for the negative sample $\mathcal{X}_{i,k}^{-}$ constructed by Gaussian Threshold Sampling method introduced in Section \ref{section_guassian_sampling}. $\tau$ is the temperature coefficient. $s$ is the scoring function. $s_{i,i}$ represents the score value between the anchor sample and the positive sample, while $s_{i,k}$ represents the score value between the anchor sample and the negative sample. Minimizing InfoNCE encourages high similarity between $\mathcal{Z}_i$ and $\mathcal{Z}_{i}^{+}$ and low similarity between $\mathcal{Z}_i$ and $\mathcal{Z}_{i,k}^{-}$, reinforcing feature distinctiveness. Ideally, the final optimization result of InfoNCE can be expressed as:
\begin{equation} \label{InfoNCE_relation_eq}
	s_{i,i}>s_{i,k}, \forall k \in [1,2,\dots,m]
\end{equation}

For the RUL prediction task in this paper, there are two reasons for improving InfoNCE:

1. Equations \ref{fsgri_eq} and \ref{fsgri_y_eq} can be further described as relationships between positive and negative sample pairs and the anchor sample:
\begin{equation} \label{fsgri_relation_eq}
	s_{i,i}>s_{i,1}>s_{i,2}>\dots>s_{i,m}
\end{equation}
\begin{equation} \label{fsgri_sample_pair_y_eq}
	\mathcal{Y}_i>\mathcal{Y}_{i,1}>\mathcal{Y}_{i,2}>\dots>\mathcal{Y}_{i,m}
\end{equation}
Standard InfoNCE tends to simultaneously minimize all terms involving $s_{i,k}$  as shown in Equation  \ref{InfoNCE_relation_eq}, which fails to satisfy the Equation \ref{fsgri_relation_eq}.

2. Analyzing the gradient of InfoNCE reveals:
\begin{align}
	\begin{split}
		\frac{\partial\mathcal{L}}{\partial\mathcal{X}_{i,k}^{-}} &= \frac{\partial\mathcal{L}}{\partial s_{i,k}}~\frac{\partial s_{i,k}}{\partial\mathcal{X}_{i,k}^{-}}\\
		&= \frac{1}{\tau}\left( \frac{{\exp\left( \frac{s_{i,k}}{\tau} \right)}\mathcal{Z}_{i}}{{\sum_{k! = i}{\exp\left( \frac{s_{i,k}}{\tau} \right)}} + {\exp\left( \frac{s_{i,i}}{\tau} \right)}} \right)\\
		&= \frac{1}{\tau}{\exp\left( \frac{s_{i,k}}{\tau} \right)}P
	\end{split}
\end{align}
where:
\begin{equation}
	P = \left( \frac{\mathcal{Z}_{i}}{{\sum_{k! = i}{\exp\left( \frac{s_{i,k}}{\tau} \right)}} + {\exp\left( \frac{s_{i,i}}{\tau} \right)}} \right)
\end{equation}
As $P$ is independent of the negative sample index, the gradient magnitude of the negative sample $\mathcal{X}_{i,k}^{-}$ during the update process is related to $s_{i,k}$. In other words, negative samples with higher similarity to $\mathcal{X}_i$  contribute more to the gradient during model updates. These samples are referred to as "hard samples"\cite{r20He2020-hc}. Optimizing hard samples to have lower similarity to $\mathcal{X}_i$ is beneficial in many tasks.

However, in RUL prediction, it's not straightforward to determine if a sample is a hard sample based solely on similarity. Samples with higher similarity might be close to $\mathcal{X}_i$ in terms of the label value (RUL), and these samples, even though they have high similarity to $\mathcal{X}_i$, are not the samples targeted for specific optimization (referred to as "false hard samples"). It's normal for the features of such samples to have high similarity to $\mathcal{X}_i$. Therefore, we need to establish a relationship between negative samples and RUL, allowing InfoNCE to ignore false hard samples and focus on optimizing real hard samples that have high similarity but are distant in terms of RUL from $\mathcal{X}_i$.

Based on the above two points, the improved Distance Weighted InfoNCE (DW-InfoNCE) is as follows:
\begin{align} \label{DW_InfoNCE}
	\begin{split}
		\begin{cases}
			\mathcal{L}_{DW - InfoNCE}~ = - {\log\left( \frac{\exp\left( \frac{s_{i,i}}{\tau} \right)}{{\sum_{k! = i}{\exp\left(\alpha_{i,k} \frac{s_{i,k}}{\tau} \right)}} + {\exp\left( \frac{s_{i,i}}{\tau} \right)}} \right)}\\
			\alpha_{i,k} = \lambda\left( {\mathcal{Y}_{i} - \mathcal{Y}_{i,k}} \right)^{2}
		\end{cases}
	\end{split}
\end{align}
where $\mathcal{Y}_i$ and $\mathcal{Y}_{i,k}$ are the RUL values for the anchor sample $\mathcal{X}_i$ and the negative sample $\mathcal{X}_{i,k}^{-}$, respectively. The distance weight $\alpha_{i,k}$ constructed based on RUL calibrates the distance of each negative sample from $\mathcal{X}_i$, and it is scaled by $\lambda$. As per Equation \ref{fsgri_sample_pair_y_eq}, $\alpha_{i,k}$ satisfies:
\begin{equation}
	\alpha_{i,i}<\alpha_{i,1}<\alpha_{i,2}<\dots<\alpha_{i,m}
\end{equation}
That is, the farther the negative sample is from $\mathcal{X}_i$ , the larger its weight $\alpha_{i,k}$, and consequently, the negative sample gradient is adjusted accordingly and increases, as indicated by Equation \ref{dw_infonce_gradient}. 
\begin{equation} \label{dw_infonce_gradient}
	\frac{\partial \mathcal{L}}{\partial \mathcal{X}_{i,k}^{-}}=\frac{\alpha_{i,k}}{\tau}{\exp\left( {\alpha_{i,k}\frac{s_{i,k}}{\tau}} \right)}\hat{P}
\end{equation}
where:
\begin{equation}
	\hat{P} = \left( \frac{\mathcal{Z}_{i}}{{\sum_{k! = i}{\exp\left(\alpha_{i,k} \frac{s_{i,k}}{\tau} \right)}} + {\exp\left( \frac{s_{i,i}}{\tau} \right)}} \right)
\end{equation}
A larger gradient implies that the model is more inclined to optimize these samples to lower similarity, thereby encouraging the model to achieve the relationship given in Equation \ref{fsgri_relation_eq}. Moreover, $\alpha_{i,k}$  significantly reduces the weight of samples closer to $\mathcal{X}_i$ , effectively mitigating the impact of false hard samples on model optimization. In summary, the distance weight $\alpha_{i,k}$ simultaneously addresses the two issues mentioned above, meeting the requirements of the FSGRI.

\subsubsection{Training Process} \label{section_train}
\begin{figure}[ht]
	\centering
	\includegraphics[scale=0.7]{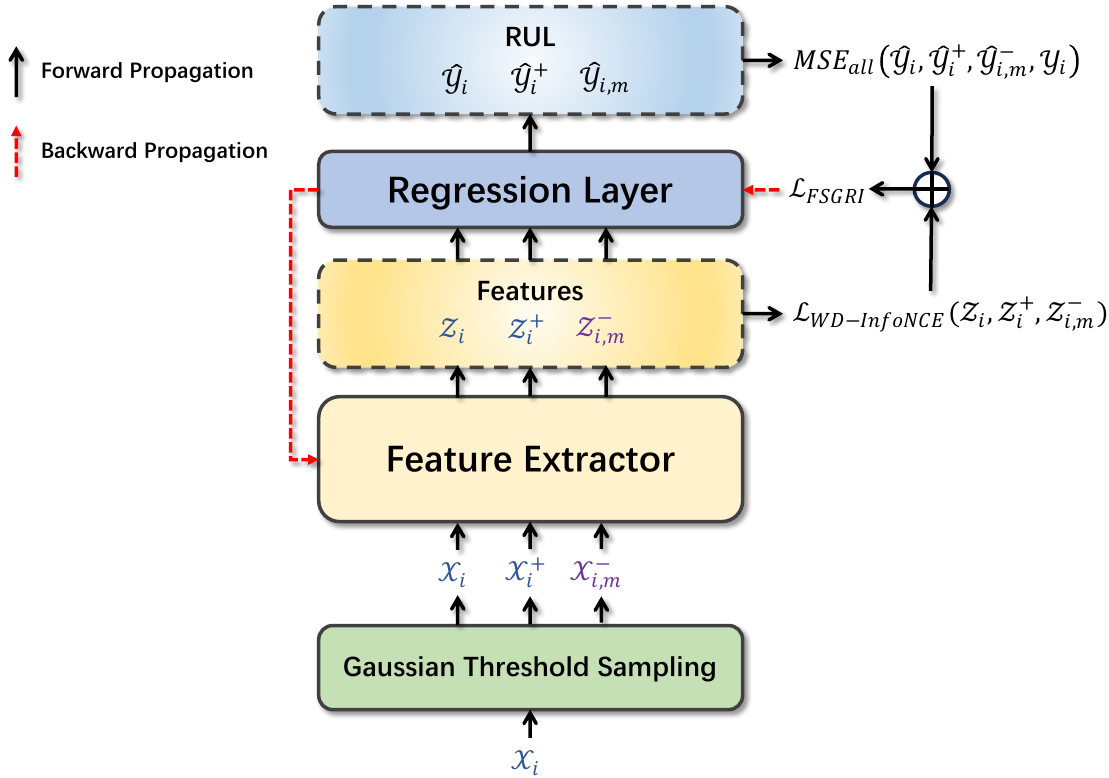}
	\caption{The illustration of proposed FSGRI training method for data-driven model.}\label{fig6}
\end{figure}

By combining the DW-InfoNCE proposed in the previous section with the Mean Squared Error (MSE) loss function used for RUL prediction, we can form the FSGRI training method proposed in this paper. The composition of the loss functions and details of FSGRI training will be elaborated below.
For most deep learning-based data-driven RUL prediction methods, the fundamental architecture can be divided into two parts: a feature extractor and a regression layer. The FSGRI training method is applicable to any model that adheres to this architecture, as illustrated in Figure \ref{fig6}.

First, after constructing positive and negative sample pairs for the sample $\mathcal{X}_i$ using the Gaussian threshold sampling method described in Section \ref{section_guassian_sampling}, high-dimensional features $\mathcal{Z}_i$, $\mathcal{Z}_{i}^{+}$ and $\mathcal{Z}_{i,m}^{+}$ are obtained through the model's feature extractor. These features are then used to calculate the DW-InfoNCE loss according to Equation \ref{DW_InfoNCE}. Subsequently, separate RUL regressions are performed for $\mathcal{Z}_i$, $\mathcal{Z}_{i}^{+}$ and $\mathcal{Z}_{i,m}^{+}$, and their respective MSE losses are calculated. The sum of these $MSE$ losses, denoted as $MSE_{all}$, is computed. The overall computation process is as follows:
\begin{equation} \label{mse_all_eq}
	MSE_{all} = MSE\left( {{\hat{\mathcal{Y}}}_{i},\mathcal{Y}_{i}} \right) + MSE\left( {{\hat{\mathcal{Y}}}_{i}^{+},\mathcal{Y}_{i}} \right) + MSE\left( {{\hat{\mathcal{Y}}}_{i,m},\mathcal{Y}_{i,m}} \right)
\end{equation} 
Ultimately, FSGRI loss function is as follows:
\begin{equation} \label{fsgri_loss_eq}
	\mathcal{L}_{FSGRI} = \mathcal{L}_{DW-InfoNCE}+MSE_{all}
\end{equation}
Through the $L_{FSGRI}$, the model weights can be optimized using gradient descent. To provide a more details for the batch gradient descent process, the pseudocode of FSGIR is presented in Algorism \ref{alg1}. 

It is noteworthy that in line \ref{line4} of Algorithm 1, the additional sampling of negative samples may result in the effective number of samples involved in the computation being larger than the specified batch size during batch gradient descent. Therefore, the batch size is scaled to ensure that the actual number of computed samples aligns with the expectations.

\begin{algorithm}
	\caption{Batch FSGRI training method.}
	\label{alg1}
	\SetAlgoLined
	\KwIn{Batch size $b$; Epoch $epoch$; Number of negative samples $m$; Feature extractor $\mathcal{F}$; Regression layer $\mathcal{A}$; Optimizer $opt$}
	\KwOut{A optimized feature extractor $\mathcal{F}$ and regression layer $\mathcal{A}$}
	Initializing all the weights in $\mathcal{F}$ and $\mathcal{A}$\\
	\For{$e < epoch$}
	{
		\While{there are samples \textbf{not} be utilized in this epoch}
		{
			Randomly Sampling a batch of samples $\mathcal{X}$ with \textbf{batch size $\lfloor b/(m+1) \rfloor$}.\label{line4}\\
			\For{$\mathcal{X}_i$ \textbf{in} $\mathcal{X}$}
			{
				Find all samples $\mathcal{X}_{i,k}$ that come from the same degradation device as $\mathcal{X}_i$;\\
				 Assign sampling probabilities to each sample in $\mathcal{X}_{i,k}$ based on Equation \ref{gaussian_prob_eq};\\
				 $\mathcal{X}_{i,m}^{-} \leftarrow$ Sampling $m$ samples from $\mathcal{X}_{i,k}$;\\
				 Constructing $\mathcal{X}_{i}^{+}$ with Equation \ref{positive_sample_eq};\\
				 $\left. \mathcal{Z}_{i}\leftarrow\mathcal{F}\left( \mathcal{X}_{i} \right),\mathcal{Z}_{i}^{+}\leftarrow\mathcal{F}\left( \mathcal{X}_{i}^{+} \right),\mathcal{Z}_{i,m}^{-}\leftarrow\mathcal{F}\left( \mathcal{X}_{i,m}^{-} \right) \right.$;\\
				 Computing $\mathcal{L}_{DW-InfoNCE}$ with Equation \ref{DW_InfoNCE};\\
				 Computing RUL $\left. \hat{\mathcal{Y}}_{i}\leftarrow\mathcal{A}\left( \mathcal{Z}_{i} \right),\hat{\mathcal{Y}}_{i}^{+}\leftarrow\mathcal{A}\left( \mathcal{Z}_{i}^{+} \right),\hat{\mathcal{Y}}_{i,m}\leftarrow\mathcal{A}\left( \mathcal{Z}_{i,m}^{-} \right) \right.$;\\
				 Computing $MSE_{all}$ with Equation \ref{mse_all_eq};\\
				 Computing $\mathcal{L}_{FSGRI}^{i}$ with Equation \ref{fsgri_loss_eq};
			}
			Computing total mean loss $\mathcal{L}_{FSGRI} \leftarrow \frac{1}{b}  {\sum\limits_{i}\mathcal{L}_{FSGRI}^{i}} $\\
			Perform gradient backpropagation and update all the weights in $F$ and $A$ using $opt$.
		}
	}
\end{algorithm}

\section{Experiments} \label{experiemnts}
\subsection{Dataset Description and Evaluation Metrics} \label{section_datasets}
This paper validates the proposed method's effectiveness using the commonly employed C-MAPSS\cite{r24Saxena2008-vr} dataset in RUL prediction. The C-MAPSS dataset, proposed by NASA Ame Prediction, consists of four sub-datasets, and their basic information is shown in Table \ref{table1}.
\begin{table*}[width=0.8\textwidth,pos=!h]
	\caption{Description of Aircraft Engine Dataset C-MAPSS} \label{table1}
	\begin{tabular*}{\tblwidth}{@{}LLLLL@{}}
		\toprule
		\pmb{Datasets} & \pmb{FD001} & \pmb{FD002} & \pmb{FD003} & \pmb{FD004} \\
		\midrule
		Number of training engines & 100 & 260 & 100 & 249 \\
		Number of test engines & 100 & 259 & 100  & 248 \\
		Operation conditions & 1 & 6 & 1  & 6 \\
		Fault modes & 1 & 1 & 2 & 2 \\
		\bottomrule
	\end{tabular*}
\end{table*}

Each sub-dataset of C-MAPSS has been divided into a training set and a test set. The training set includes operational data throughout the entire lifecycle of multiple engines, while the test set data is incomplete and randomly truncated at some point before engine failure. Lifecycle data consists of monitoring values from multiple sensors, recording parameters during each engine run. There are 21 variables, as shown in Table \ref{table2}.
\begin{table*}[width=0.5\textwidth,pos=!h]
	\caption{All availabel monitoring varibles in C-MAPSS dataset.} \label{table2}
	\begin{tabular*}{\tblwidth}{@{}CCC@{}}
		\toprule
		\textbf{Index}	&\textbf{Symbol}	&\textbf{Description}\\
		\midrule
		1	&T2	&Total temperature at fan inlet\\
		2	&T24	&Total temperature at LPC outlet\\
		3	&T30	&Total temperature at HPC outlet\\
		4	&T50	&Total temperature at LPT outlet\\
		5	&P2	&Presure at fan inlet\\
		6	&P15	&Total presure in bypass-duct\\
		7	&P30	&Total presure at HPC outlet\\
		8	&Nf	&Physical fan speed \\
		9	&Nc	&Physical core speed\\
		10	&Epr	&Engine pressure ratio (P50/P2)\\
		11	&Ps30	&Static presure at HPC outlet\\
		12	&Phi	&Ratio of fuel flow to Ps30\\
		13	&NRf	&Corrected fan speed\\
		14	&NRc	&Corrected core speed\\
		15	&BPR	&Bypass Ratio\\
		16	&FarB	&Burner fuel-air ratio\\
		17	&htBleed	&Bleed Enthalpy\\
		18	&Nf\_dmd	&Demanded fan speed\\
		19	&PCNfr\_dmd	&Demanded corrected fan speed\\
		20	&W31	&HPT coolant bleed\\
		21	&W32	&LPT coolant bleed\\
		\bottomrule
	\end{tabular*}
\end{table*}

This paper employs commonly used metrics in regression problems and RUL prediction, namely RMSE (Root Mean Square Error) and MAPE (Mean Absolute Percentage Error), to assess the performance of the model. The mathematical formulations of these metrics are as follows:
\begin{equation}
	RMSE\left( {\mathcal{Y},\hat{\mathcal{Y}}} \right) = \sqrt{\frac{1}{N}{\sum\limits_{i = 1}^{N}\left( {\mathcal{Y}_{i} - {\hat{\mathcal{Y}}}_{i}} \right)^{2}}}
\end{equation}
\begin{equation}
	MAPE\left( {\mathcal{Y},\hat{\mathcal{Y}}} \right) = \frac{100\%}{N}{\sum\limits_{i = 1}^{N}\left| \frac{\mathcal{Y}_{i} - {\hat{\mathcal{Y}}}_{i}}{\mathcal{Y}_{i}} \right|}
\end{equation}
where N is the number of samples, $\mathcal{Y}_i$ is the true RUL of the $i$-th sample, and $\hat{\mathcal{Y}}_i$  is the predicted RUL

\subsection{Dataset Preprocessing}
The C-MAPSS dataset contains some variables with constant values that do not provide meaningful information. Therefore, these variables are removed and excluded from the model training and testing processes. Following the approach in the literature\cite{r6Zhang2023-kc}, corresponding to the indices in Table \ref{table2}, the selected variable indices are 2, 3, 4, 7, 8, 9, 11, 12, 13, 14, 15, 17, 20, and 21.

\begin{figure}[ht]
	\centering
	\includegraphics[scale=0.7]{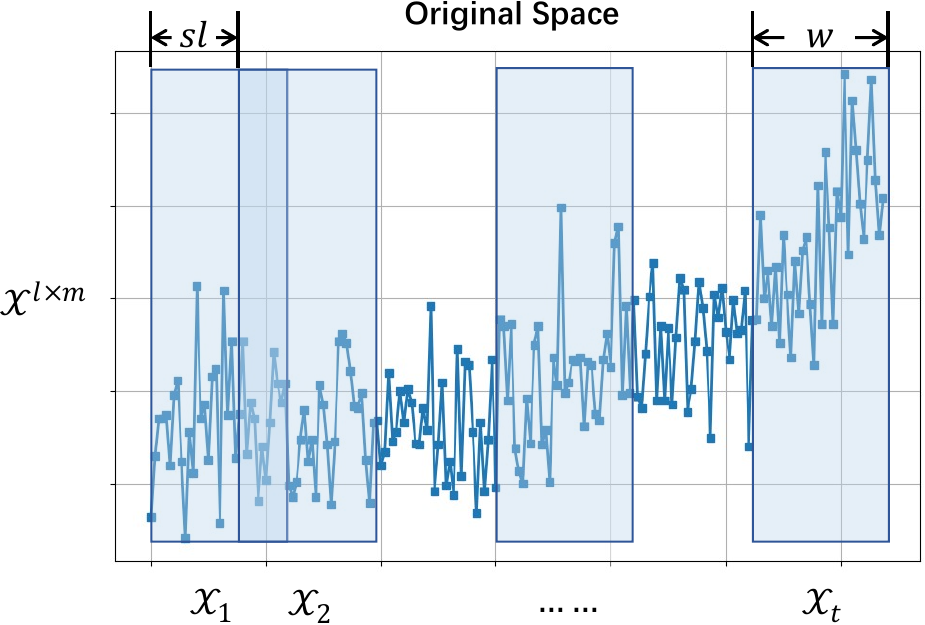}
	\caption{ Illustration of Sliding window method used for constructing input samples.}\label{fig7}
\end{figure}

To construct input samples for the data-driven model, the sliding window method is a commonly used approach. As shown in Figure \ref{fig7}, for the original time series data with a length of $l$ composed of $m$ sensors, a window of size $w$ is slid along the time dimension with a step size of $sl$. The data within each window is considered as one sample, and each sample $\mathcal{X}_i \in \mathbb{R}^{w \times m}$.

Data normalization is a standard procedure in data-driven methods. In this paper, the Min-Max normalization method is utilized to normalize all samples. Its formulation is as follows:
\begin{equation}
	{\overline{{\mathcal{X}}}}_{i}^{w \times m} = \frac{\mathcal{X}_{i}^{w \times m} - \mathcal{X}_{min}^{1 \times m}}{\mathcal{X}_{max}^{1 \times m} - \mathcal{X}_{min}^{1 \times m}}
\end{equation}
where $\mathcal{X}_{min}^{1\times m}$ and $\mathcal{X}_{max}^{1\times m}$ are the minimum and maximum values of the data from each of the $m$ sensors, and $\mathcal{X}_{i}^{w \times m}$  is the $i$-th original data sample. 

Segmented linear degradation labels are a commonly employed labeling method for RUL in C-MAPSS data\cite{r5Zhang2023-xb}\cite{r6Zhang2023-kc}\cite{r7Xu2023-cl}\cite{r11Zhang2024-aw}. The construction process is as follows:
\begin{align}
	\begin{cases}
		\mathcal{Y}_k = 1, &k\geq125\\
		\mathcal{Y}_k = \frac{k}{l-125}, &k<125 ~~and~~ l> 125\\
		\mathcal{Y}_k = \frac{k}{l}, &l<125\\
	\end{cases}
\end{align}
where $\mathcal{Y}_i$ is the RUL percentage corresponding to the $k$-th cycle, and $l$ is the total number of cycles. After constructing samples using the sliding window method, the RUL of the last cycle within the sample window is taken as the label for each sample, which serves as the prediction target for the model. 

\subsection{Tuning Experiments} \label{tuning_experiments}
\begin{figure}[ht]
	\centering
	\subfigure[FD001]{
		\includegraphics[scale=0.45]{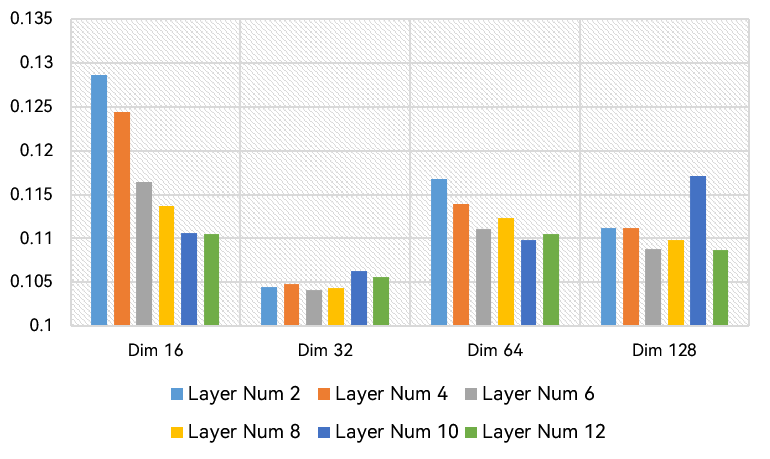}
	}
	\subfigure[FD003]{
	\includegraphics[scale=0.45]{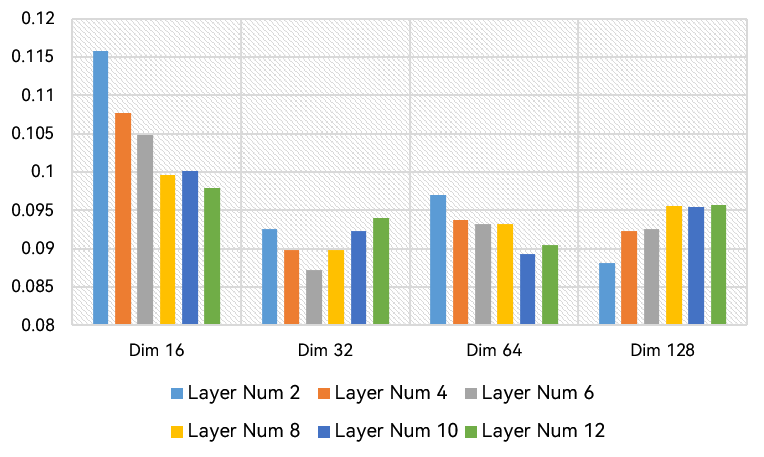}
	}
	\subfigure[FD002]{
	\includegraphics[scale=0.45]{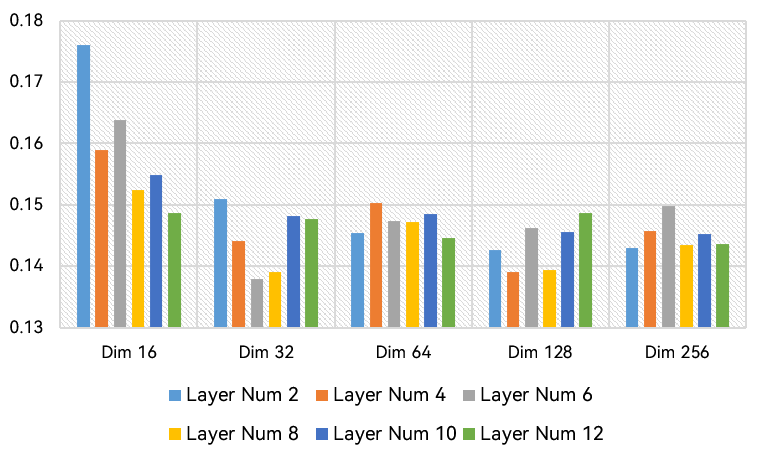}
	}
	\subfigure[FD004]{
	\includegraphics[scale=0.45]{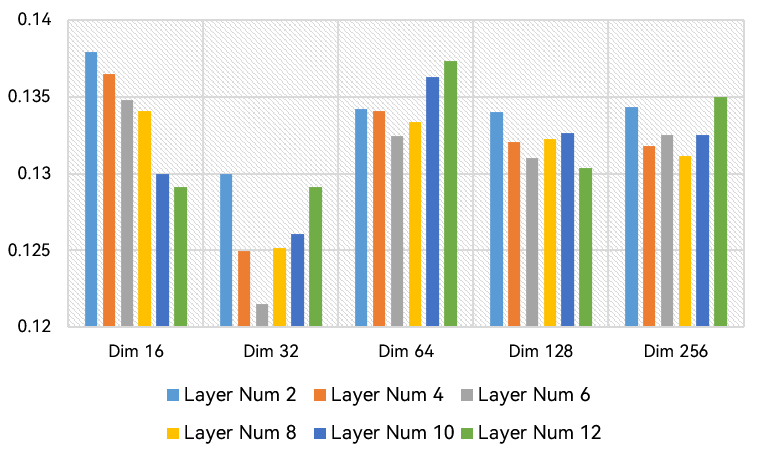}
	}
	\caption{The RMSE obtained by Dual-Mixer with different hyperparameters on the four datasets.}\label{fig8}
\end{figure}

In this section, the experiment will first determine the two main hyperparameters of the proposed Dual-Mixer: the number of DML layers and the feature dimension d. At the same time, the effectiveness of the model will be preliminarily verified. The number of DML layers is adjusted within the range [2, 4, 6, 8, 10, 12], and the feature dimension is adjusted within the range [16, 32, 64, 128] (for FD001 and FD003) and [16, 32, 64, 128, 256] (for FD002 and FD004). Using a pairwise combination approach, validation is conducted on the four sub-datasets, and the experimental results are shown in Figure \ref{fig8}.
In Figure \ref{fig8}, each subplot's X-axis represents the model's feature dimension. For FD001 and FD003, due to the smaller data size, feature dimensions are selected from [16, 32, 64, 128]. For FD002 and FD004, considering the larger data size, an additional dimension of 256 is considered, and feature dimensions are selected from [16, 32, 64, 128, 256]. The color of each bar in the plot represents the number of DML layers, with lighter colors indicating more layers. The Y-axis represents the RMSE value, where lower values indicate higher predictive accuracy.

From the figure, it can be observed that, in most cases, a feature dimension of 32 and 6 layers of DML are most suitable, achieving the best RMSE in the majority of situations. Therefore, for this experiment, a feature dimension d of 32 and 6 DML layers are chosen. The performance with a feature dimension of 16 is generally worse, indicating insufficient non-linear mapping capability and occurrence of underfitting. Increasing the number of DML layers significantly improves the model's performance at a feature dimension of 16. As the feature dimension increases and the number of DML layers deepens, there is an increasing risk of overfitting. However, due to the residual connections and LayerNorm within DML, the overfitting phenomenon is not very pronounced, and the network's performance does not significantly degrade as the number of layers increases. Therefore, when applying Dual-Mixer to other datasets, careful adjustment of the feature dimension is recommended. 

\subsection{Comparison Experiments}
To further validate the effectiveness of the proposed Dual-Mixer model and the FSGRI training method, we selected the following state-of-the-art models as baseline methods:

1) IMDSSN\cite{r5Zhang2023-xb}: A recent RUL prediction method based on Transformer encoder and attention mechanisms.

2) BTSAM\cite{r6Zhang2023-kc}: A RUL prediction model based on bidirectional GRUs and temporal attention mechanisms.

3) CNN-GRU\cite{r6Zhang2023-kc}: A multi-dimensional feature fusion network using convolutional networks and GRUs for RUL prediction.

4) DAMCNN\cite{r25Jiang2022-ib}: A convolutional neural network incorporating multiple attention mechanisms for RUL prediction.

5) MLP-Mixer\cite{r15Tolstikhin2021-xm}: A deep network architecture for extracting multi-dimensional features, applicable to time series data. In our experiments, a 2-layer MLP-Mixer with 128 feature dimensions was used.

6) TS-Mixer\cite{r14Ekambaram2023-bk}: A recent time series prediction model based on improvements to MLP-Mixer. It exhibits excellent performance in multivariate time series prediction and is adapted for RUL prediction. In our experiments, we used the same configuration as Dual-Mixer, i.e., 6 layers with 32 feature dimensions.

7) LSTM\cite{LSTM_Base}: A basic deep learning method for processing time series data. We used a 3-layer LSTM with 256 dimensions.

We reproduced these methods in the experiments, and the code can be found at:\url{https://github.com/fuen1590/PhmDeepLearningProjects}. All these methods follow the architecture shown in Figure \ref{fig6}, making them compatible with the FSGRI training method.
The hyper-parameters configuration for the proposed Dual-Mixer method is provided in Table \ref{table3}.
\begin{table*}[width=0.8\textwidth]
	\caption{Experimental hyper-parameters configuration.} \label{table3}
	\begin{tabular*}{\tblwidth}{CCp{9cm}}
		\toprule
		\textbf{Params}	&\textbf{Value}	&\textbf{Description}\\
		\midrule
		$b$		&128		&Bach size\\
		$lr$		&1e-2		&Learning rate.\\
		$w$		&30		&Sliding window size. \\
		$sl$		&1		&Sliding window stride.\\
		$opt$		&Adam		&The backpropgation optimizer. \\
		$epoch$		&100		&The maxium training epoch.\\
		$N$		&6		&The number of DML layers.\\
		$d$		&32		&The feature dimension of Dual-Mixer model.\\
		$m$		&5		&The number of negative samples.\\
		$\beta$		&0.4		&Threshold coefficient for Gaussian threshold sampling.\\
		$\sigma_1$		&0.3		&Gaussian threshold sampling standard deviation.\\
		$\sigma_2$		&0.15		&The standard deviation of the Gaussian noise used when constructing positive samples.\\
		$\lambda$		&2.0		&Coefficient of $\alpha_k$  in DW-InfoNCE.\\
		\bottomrule
	\end{tabular*}
\end{table*}

All experiments in this study were implemented on Ubuntu 18.02 with PyTorch 2.0. The inference and training were accelerated using Nvidia GeForce RTX 3090. The final experimental results are summarized in Table \ref{table4}.
\begin{table*}[width=0.8\textwidth]
	\caption{Comparison Experiments Results in C-MAPSS dataset.} \label{table4}
	\begin{tabular*}{\tblwidth}{CCCCCCCCC} 
		\toprule
		\multirow{2}*{\textbf{Models}}	&	\multicolumn{2}{c}{\textbf{FD001}}	&	\multicolumn{2}{c}{\textbf{FD002}}	&	\multicolumn{2}{c}{\textbf{FD003}}	&	\multicolumn{2}{c}{\textbf{FD004}} \\
		~	&	RMSE	&	MAPE	&	RMSE	&	MAPE	&	RMSE	&	MAPE	&	RMSE	&	MAPE \\
		\toprule
		LSTM	&	0.1279	&	13.52\%	&	0.1747	&	20.74\%	&	0.1116	&	10.63\%	&	0.1542	&	16.16\%\\
		LSTM-F	&	\underline{0.1098}	&	\underline{10.91\%}	&	\underline{0.1460}	&	\underline{17.12\%}	&	\underline{0.0938}	&	\underline{7.82\%}	&	\underline{0.1500}	&	\underline{15.86\%}\\
		Improvement	&	14.15\%	&	2.61\%	&	16.42\%	&	3.62\%	&	15.96\%	&	2.81\%	&	2.77\%	&	0.31\%\\
		\midrule
		CNN-GRU	&	0.1166	&	12.52\%	&	0.1493	&	17.30\%	&	0.0985	&	9.28\%	&	0.1486	&	15.68\%\\
		CNN-GRU-F	&	\underline{0.1080}	&	\underline{11.24\%}	&	\underline{0.1477}	&	\underline{17.19\%}	&	\underline{0.0968}	&	\underline{8.79\%}	&	\underline{0.1355}	&	\underline{13.62\%}\\
		Improvement	&	7.38\%	&	1.28\%	&	1.08\%	&	0.11\%	&	1.77\%	&	0.49\%	&	8.79\%	&	2.06\%\\
		\midrule
		DAMCNN	&	\underline{0.1174}	&	12.29\%	&	0.1960	&	25.45\%	&	\underline{0.0961}	&	\underline{8.29\%}	&	0.1767	&	19.10\%\\
		DAMCNN-F	&	0.1192	&	\underline{11.21\%}	&	\underline{0.1876}	&	\underline{21.93\%}	&	0.1075	&	10.52\%	&	\underline{0.1755}	&	\underline{16.38\%}\\
		Improvement	&	-1.53\%	&	1.08\%	&	4.29\%	&	3.52\%	&	-11.86\%	&	-2.23\%	&	0.063\%	&	2.71\%\\
		\midrule
		BTSAM	&	0.1062	&	9.99\%	&	\underline{0.1418}	&	18.54\%	&	0.0925	&	10.39\%	&	0.1429	&	12.26\%\\
		BTSAM-F	&	\underline{0.1037}	&	\underline{9.81\%}	&	0.1475	&	\underline{16.23\%}	&	\underline{\textbf{0.0887}}	&	\underline{7.77\%}	&	\underline{0.1376}	&	\underline{12.19\%} \\
		Improvement	&	2.35\%	&	0.18\%	&	-4.02\%	&	2.31\%	&	4.60\%	&	2.63\%	&	3.69\%	&	0.55\%\\
		\midrule
		IMDSSN	&	0.1100	&	11.39\%	&	\underline{0.1355}	&	\underline{16.51\%}	&	0.0959	&	9.67\%	&	\underline{0.1273}	&	\underline{13.98\%}\\
		IMDSSN-F	&	\underline{0.1044}	&	\underline{9.87\%}	&	0.1368	&	17.16\%	&	\underline{0.0903}	&	\underline{8.04\%}	&	0.1278	&	16.13\%\\
		Improvement	&	5.09\%	&	1.52\%	&	-0.96\%	&	-0.65\%	&	5.82\%	&	1.64\%	&	-0.34\%	&	-2.15\%\\
		\midrule
		MLP-Mixer	&	0.1837	&	23.98\%	&	0.1746	&	21.39\%	&	0.1480	&	17.30\%	&	0.1534	&	14.83\%\\
		MLP-Mixer-F	&	\underline{0.1115}	&	\underline{12.55\%}	&	\underline{0.1371}	&	\underline{15.11\%}	&	\underline{0.1312}	&	\underline{15.21\%}	&	\underline{0.1269}	&	\underline{12.97\%}\\
		Improvement	&	39.30\%	&	11.43\%	&	21.48\%	&	6.28\%	&	11.35\%	&	2.09\%	&	17.28\%	&	1.86\%\\
		\midrule
		TS-Mixer	&	0.1161	&	12.92\%	&	0.1911	&	44.50\%	&	0.1028	&	9.84\%	&	0.1629	&	\underline{16.56\%}\\
		TS-Mixer-F	&	\underline{0.1042}	&	\underline{10.85\%}	&	\underline{0.1415}	&	\underline{16.54\%}	&	\underline{0.0910}	&	\underline{8.43\%}	&	\underline{0.1436}	&	19.91\%\\
		Improvement	&	10.25\%	&	2.07\%	&	25.96\%	&	27.96\%	&	11.45\%	&	1.40\%	&	11.85\%	&	-3.35\%\\
		\midrule
		\textbf{Dual-Mixer (Ours)}	&	0.1041	&	10.18\%	&	0.1390	&	14.98\%	&	\textbf{\underline{0.0887}}	&	\textbf{\underline{7.65\%}}	&	0.1215	&	12.36\%\\
		\textbf{Dual-Mixer-F}	&	\textbf{\underline{0.1002}}	&	\textbf{\underline{9.21\%}}	&	\textbf{\underline{0.1338}}	&	\textbf{\underline{13.42\%}}	&	\textbf{\underline{0.0887}}	&	7.79\%	&	\textbf{\underline{0.1141}}	&	\textbf{\underline{10.79\%}}\\
		Improvement	&	3.75\%	&	0.97\%	&	3.74\%	&	1.56\%	&	0\%	&	-0.13\%	&	6.11\%	&	1.56\%\\
		\bottomrule
	\end{tabular*}
\end{table*}

The models in the table with the suffix "-F" represent the results obtained using FSGRI training, while models without the suffix were trained using conventional gradient descent. The underscore (\_) indicates improved performance metrics compared to the original method when using FSGRI, and bold font represents the globally optimal performance metric. The "Improvement" in the table indicates the extent to which FSGRI enhances the model's performance. All experimental results are based on the average of three trials.

We compared each model using normal training methods with FSGRI training methods. In almost all cases, FSGRI consistently led to stable improvements in predictive performance. For all models using FSGRI training, the improvement rates for RMSE and MAPE were, on average, 9.14\% and 2.38\% in FD001, 7.55\% and 4.97\% in FD002, 4.34\% and 0.97\% in FD003, and 6.98\% and 1.32\% in FD004. Overall, FSGRI leads to average improvements of 7.00\% and 2.41\% in RMSE and MAPE, respectively, across all models in the C-MAPSS dataset.
\begin{figure}
	\centering
		\includegraphics[scale=0.14]{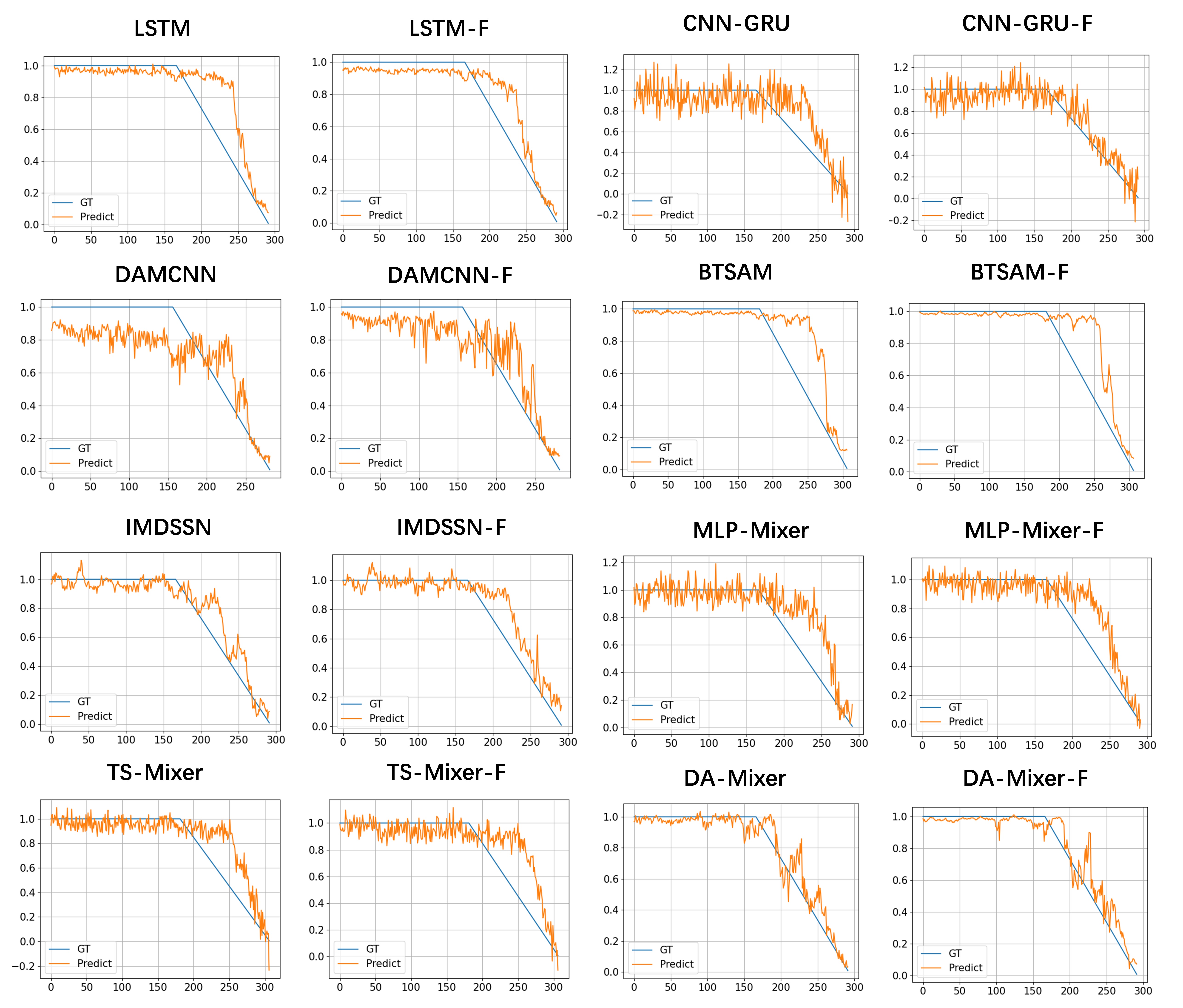}
	\caption{The RUL prediction results of Engine 1 in FD004 test dataset.}\label{fig9}
\end{figure}

In cases without FSGRI, Dual-Mixer, except for a slightly higher MAPE in FD001 and FD004 compared to BTSAM, achieved the optimal values in all metrics among all models. For an intuitive assessment of the models' RUL prediction capabilities, Figure \ref{fig9} presents the prediction results for all models on the FD004 test set for Engine 1. Compared to other models, the proposed Dual-Mixer demonstrates a stronger ability to capture detailed features in the data. For example, around 200 cycles, Dual-Mixer maintains good predictive capabilities.

\subsection{Features Visualization}
\begin{figure}
	\centering
	\includegraphics[scale=0.14]{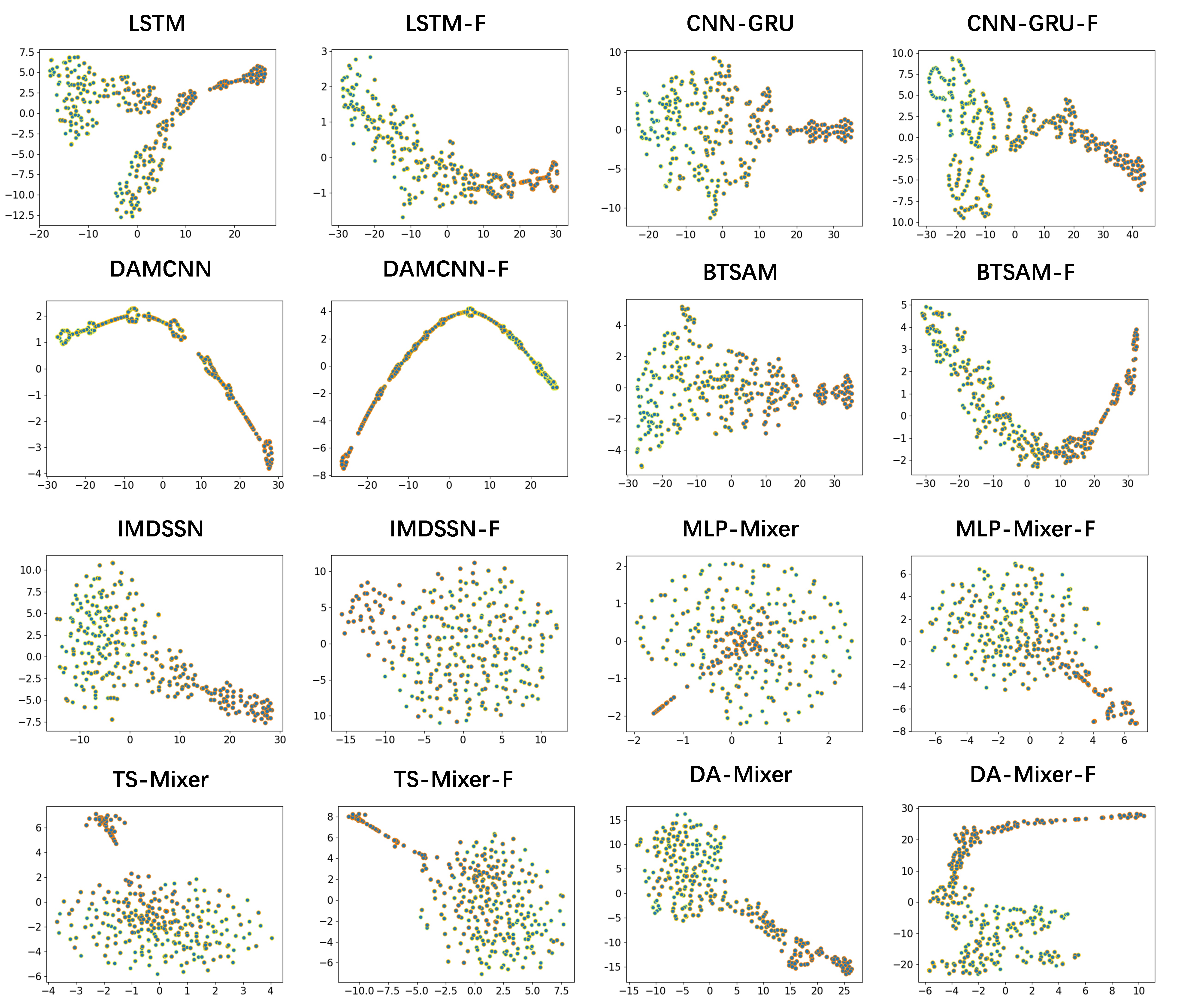}
	\caption{The RUL prediction results of Engine 1 in FD004 test dataset.}\label{fig10}
\end{figure}

We visualized the output features of the models before and after FSGRI training to further analyze the impact of FSGRI on the distribution of model features, as shown in Figure \ref{fig10}. In the visualization process, we selected data from Engine 1 of the FD004 test set as the test sample and applied dimensionality reduction and visualization using the t-Distributed Stochastic Neighbor Embedding (t-SNE)[28] method to the features of all models before and after FSGRI.
In the figure, each green point represents the features of an input sample. The color intensity at the edge of the green point indicates the corresponding sample's RUL, with darker colors representing lower RUL (indicating that the sample is in the later stages of degradation).
From the Figure \ref{fig10}, it can be observed that after using FSGRI, the feature distribution tends to become smoother, particularly in the middle to later stages of degradation, where sample features become more pronounced. This is reflected in the feature distribution becoming more discriminative. For instance, in the case of Dual-Mixer proposed in this paper, after FSGRI training, the features in the later stages exhibit a clear continuous distribution, making them more distinctive compared to the feature distribution before FSGRI training. This phenomenon is also evident in the BTSAM, CNN-GRU, and TS-Mixer models. Thus, FSGRI has a noticeable smoothing effect on the feature distribution for most models.

\subsection{Ablation Study}
In order to understand the importance of each component in Dual-Mixer, five variant models were designed as follows:\\
1) Dual-Mixer-oCm: This variant removes the Gate Block used for dual-path feature exchange in DML, retaining only the Gate Block in the feature output section.\\
2) Dual-Mixer-oCO: This variant simultaneously removes the Gate Block used for dual-path feature exchange and the Gate Block in the output section.\\
3) Dual-Mixer-oO: This variant only removes the Gate Block in the output section.\\
4) Dual-Mixer-oT: This variant removes the Spatial Part while keeping the rest unchanged.\\
5) Dual-Mixer-oS: This variant removes the Temporal Part while keeping the rest unchanged.\\

By designing these five variant models, the goal is to analyze the rationality of our design and whether the model effectively utilizes multi-path features. Comparative experiments were conducted on four datasets, as shown in Table \ref{table5}.
\begin{table*}[width=0.8\textwidth]
	\caption{Ablation study results.} \label{table5}
	\begin{tabular*}{\tblwidth}{CCCCCCCCC} 
		\toprule
		\multirow{2}*{\textbf{Models}}	&	\multicolumn{2}{c}{\textbf{FD001}}	&	\multicolumn{2}{c}{\textbf{FD002}}	&	\multicolumn{2}{c}{\textbf{FD003}}	&	\multicolumn{2}{c}{\textbf{FD004}} \\
		~	&	RMSE	&	MAPE	&	RMSE	&	MAPE	&	RMSE	&	MAPE	&	RMSE	&	MAPE \\
		\midrule
		Dual-Mixer	&	\textbf{0.1041}	&	\textbf{10.18\%}	&	\textbf{0.1390}	&	\textbf{14.98\%}	&	\textbf{0.0887}	&	\textbf{7.65\%}	&	\textbf{0.1215}	&	\textbf{12.36\%} \\
		Dual-Mixer-oCm	&	0.1049	&	14.19\%	&	0.1408	&	17.01\%	&	0.0893	&	7.88\%	&	0.1341	&	16.46\% \\
		Dual-Mixer-oCO	&	0.1061	&	10.73\%	&	0.1435	&	19.54\%	&	0.0893	&	7.87\%	&	0.1412	&	21.12\%\\
		Dual-Mixer-oO	&	0.1063	&	10.50\%	&	0.1392	&	18.77\%	&	0.0892	&	7.78\%	&	0.1222	&	13.72\%\\
		Dual-Mixer-oT	&	0.1092	&	12.57\%	&	0.1442	&	20.43\%	&	0.0939	&	9.02\%	&	0.1245	&	15.10\%\\
		Dual-Mixer-oS	&	0.1067	&	10.44\%	&	0.1410	&	15.96\%	&	0.0950	&	7.74\%	&	0.1484	&	15.36\%\\
		\bottomrule
	\end{tabular*}
\end{table*}

The variant Dual-Mixer-oO, which removes the output gate, experiences a relatively minor performance decline. This is because feature exchange occurs at each layer, and the role of the Gate Block in the output section is relatively small. In contrast, the performance of Dual-Mixer-oCm, which removes the Gate Block used for internal feature exchange in DML, is more severe compared to Dual-Mixer-oO. This indicates the importance of the Gate Block for internal feature exchange in DML. The most significant performance drop is observed in Dual-Mixer-oT, which removes the Spatial Part, handling only temporal features. This suggests that spatial relationships among variables play a crucial role in the RUL prediction task proposed in this paper. The experimental results demonstrate that each module of Dual-Mixer has different levels of importance and is indispensable for the model's overall performance.

\section{Conclusion} \label{conclusion}
This paper has introduced the Dual-Mixer, a flexible feature fusion model with a spatial-temporal homogeneous feature extractor, enhancing feature fusion methods in RUL prediction and improving predictive performance. Additionally, FSGRI constraint has been proposed to smooth the feature space distribution in deep learning-based data-driven RUL methods. The relationships between samples feature has been aligned with the degradation process of the device by FSGRI, simplifying the regression task in RUL prediction. This paper has presented a specific general method for implementing FSGRI based on contrastive learning, along with a Gaussian threshold sampling method and an improved DW-InfoNCE loss. Finally, the effectiveness of the proposed Dual-Mixer and FSGRI training method has been validated through comparisons with state-of-the-art RUL prediction methods on the C-MAPSS dataset. The Dual-Mixer has demonstrated superior performance across nearly all metrics, and the proposed FSGRI training method has led to an average improvement of 7.00\% and 2.41\% in RMSE and MAPE metrics for each model. The combination of Dual-Mixer and FSGRI training method yielded the optimal performance metrics among the compared methods.

\section*{Acknowledgment}
This work is supported by the National Natural Science Foundation of China under Grants 62273038 and U21A20483.

% Uncomment and use as the case may be
%\begin{theorem} 
%\end{theorem}

% Uncomment and use as the case may be
%\begin{lemma} 
%\end{lemma}


\printcredits

%\bibliographystyle{cas-model2-names}
\bibliographystyle{unsrt}

\bibliography{cas-refs}

\begin{thebibliography}{10}

\bibitem{r1Compare2017-pe}
Michele Compare, Luca Bellani, and Enrico Zio.
\newblock Reliability model of a component equipped with {PHM} capabilities.
\newblock {\em Reliab. Eng. Syst. Saf.}, 168:4--11, 2017.

\bibitem{Wang2023-sm}
Shuhui Wang, Yaguo Lei, Bin Yang, Xiang Li, Yue Shu, and Na~Lu.
\newblock A graph neural network-based data cleaning method to prevent
  intelligent fault diagnosis from data contamination.
\newblock {\em Eng. Appl. Artif. Intell.}, 126(107071):107071, 2023.

\bibitem{Wang2022-da}
Menghui Wang, Xin Ma, Yu~Hu, and Youqing Wang.
\newblock Gear fault diagnosis based on variational modal decomposition and
  wide+narrow visual field neural networks.
\newblock {\em IEEE Trans. Autom. Sci. Eng.}, 19(4):3288--3299, 2022.

\bibitem{r2Han2021-tg}
Xiao Han, Zili Wang, Min Xie, Yihai He, Yao Li, and Wenzhuo Wang.
\newblock Remaining useful life prediction and predictive maintenance
  strategies for multi-state manufacturing systems considering functional
  dependence.
\newblock {\em Reliab. Eng. Syst. Saf.}, 210(107560):107560, 2021.

\bibitem{Li2023-ck}
Xiwei Li, Yaguo Lei, Mingzhong Xu, Naipeng Li, Dengke Qiang, Qubing Ren, and
  Xiang Li.
\newblock A spectral self-focusing fault diagnosis method for automotive
  transmissions under gear-shifting conditions.
\newblock {\em Mech. Syst. Signal Process.}, 200(110499):110499, 2023.

\bibitem{r3Li2022-ou}
Xiang Li, Yuchen Jiang, Yiliu Liu, Jiusi Zhang, Shen Yin, and Hao Luo.
\newblock {RAGCN}: Region aggregation graph convolutional network for bone age
  assessment from x-ray images.
\newblock {\em IEEE Trans. Instrum. Meas.}, 71:1--12, 2022.

\bibitem{r4Zhang2022-zn}
Jiusi Zhang, Yuchen Jiang, Shimeng Wu, Xiang Li, Hao Luo, and Shen Yin.
\newblock Prediction of remaining useful life based on bidirectional gated
  recurrent unit with temporal self-attention mechanism.
\newblock {\em Reliab. Eng. Syst. Saf.}, 221(108297):108297, 2022.

\bibitem{r5Zhang2023-xb}
Jiusi Zhang, Xiang Li, Jilun Tian, Hao Luo, and Shen Yin.
\newblock An integrated multi-head dual sparse self-attention network for
  remaining useful life prediction.
\newblock {\em Reliab. Eng. Syst. Saf.}, 233(109096):109096, 2023.

\bibitem{r6Zhang2023-kc}
Jiusi Zhang, Jilun Tian, Minglei Li, Jose~Ignaclo Leon, Leopoldo~Garcia
  Franquelo, Hao Luo, and Shen Yin.
\newblock A parallel hybrid neural network with integration of spatial and
  temporal features for remaining useful life prediction in prognostics.
\newblock {\em IEEE Trans. Instrum. Meas.}, 72:1--12, 2023.

\bibitem{r7Xu2023-cl}
Weiyang Xu, Quansheng Jiang, Yehu Shen, Qixin Zhu, and Fengyu Xu.
\newblock New {RUL} prediction method for rotating machinery via data feature
  distribution and spatial attention residual network.
\newblock {\em IEEE Trans. Instrum. Meas.}, 72:1--9, 2023.

\bibitem{Wang2022-lj}
Jianguo Wang, Shude Zhang, Chenyu Li, Lifeng Wu, and Yingzhou Wang.
\newblock A data-driven method with mode decomposition mechanism for remaining
  useful life prediction of lithium-ion batteries.
\newblock {\em IEEE Trans. Power Electron.}, 37(11):13684--13695, 2022.

\bibitem{r8Sateesh_Babu2016-cs}
Giduthuri Sateesh~Babu, Peilin Zhao, and Xiao-Li Li.
\newblock Deep convolutional neural network based regression approach for
  estimation of remaining useful life.
\newblock In {\em Database Systems for Advanced Applications}, pages 214--228.
  Springer International Publishing, Cham, 2016.

\bibitem{r9Ren2021-yv}
Lei Ren, Jiabao Dong, Xiaokang Wang, Zihao Meng, Li~Zhao, and M~Jamal Deen.
\newblock A data-driven {auto-CNN-LSTM} prediction model for lithium-ion
  battery remaining useful life.
\newblock {\em IEEE Trans. Industr. Inform.}, 17(5):3478--3487, 2021.

\bibitem{Li2023-hb}
Xiwei Li, Yaguo Lei, Xiang Li, and Bin Yang.
\newblock A robust wavelet-integrated residual network for fault diagnosis of
  machines with adversarial training.
\newblock In {\em 2023 {IEEE/ASME} International Conference on Advanced
  Intelligent Mechatronics ({AIM})}. IEEE, 2023.

\bibitem{Pei2023-ev}
Hong Pei, Xiaosheng Si, Tianmei Li, Zhengxin Zhang, and Yaguo Lei.
\newblock Interactive prognosis framework between deep learning and a
  stochastic process model for remaining useful life prediction.
\newblock {\em IEEE Trans. Neural Netw. Learn. Syst.}, PP:1--13, 2023.

\bibitem{r10Li2022-zp}
Xinyao Li, Jingjing Li, Lin Zuo, Lei Zhu, and Heng~Tao Shen.
\newblock Domain adaptive remaining useful life prediction with transformer.
\newblock {\em IEEE Trans. Instrum. Meas.}, 71:1--13, 2022.

\bibitem{r11Zhang2024-aw}
Yuru Zhang, Chun Su, Jiajun Wu, Hao Liu, and Mingjiang Xie.
\newblock Trend-augmented and temporal-featured transformer network with
  multi-sensor signals for remaining useful life prediction.
\newblock {\em Reliab. Eng. Syst. Saf.}, 241(109662):109662, 2024.

\bibitem{r12Zhang2022-cu}
Tianping Zhang, Yizhuo Zhang, Wei Cao, Jiang Bian, Xiaohan Yi, Shun Zheng, and
  Jian Li.
\newblock Less is more: Fast multivariate time series forecasting with light
  sampling-oriented {MLP} structures.
\newblock 2022.

\bibitem{r13Zeng2023-ft}
Ailing Zeng, Muxi Chen, Lei Zhang, and Qiang Xu.
\newblock Are transformers effective for time series forecasting?
\newblock {\em Proc. Conf. AAAI Artif. Intell.}, 37(9):11121--11128, 2023.

\bibitem{r14Ekambaram2023-bk}
Vijay Ekambaram, Arindam Jati, Nam Nguyen, Phanwadee Sinthong, and Jayant
  Kalagnanam.
\newblock {TSMixer}: Lightweight {MLP-mixer} model for multivariate time series
  forecasting.
\newblock In {\em Proceedings of the 29th {ACM} {SIGKDD} Conference on
  Knowledge Discovery and Data Mining}, New York, NY, USA, 2023. ACM.

\bibitem{r15Tolstikhin2021-xm}
Ilya Tolstikhin, Neil Houlsby, Alexander Kolesnikov, Lucas Beyer, Xiaohua Zhai,
  Thomas Unterthiner, Jessica Yung, Andreas Steiner, Daniel Keysers, Jakob
  Uszkoreit, Mario Lucic, and Alexey Dosovitskiy.
\newblock {MLP-Mixer}: An {all-MLP} architecture for vision.
\newblock 2021.

\bibitem{r16Chen2023-ut}
Si-An Chen, Chun-Liang Li, Nate Yoder, Sercan~O Arik, and Tomas Pfister.
\newblock {TSMixer}: An {all-MLP} architecture for time series forecasting.
\newblock 2023.

\bibitem{r17Hendrycks2016-dc}
Dan Hendrycks and Kevin Gimpel.
\newblock Gaussian error linear units ({GELUs}).
\newblock 2016.

\bibitem{r18Hochreiter1997-ok}
Sepp Hochreiter and J{\"u}rgen Schmidhuber.
\newblock Long short-term memory.
\newblock {\em Neural Comput.}, 9(8):1735--1780, 1997.

\bibitem{r19Xu2019-vs}
Jingjing Xu, Xu~Sun, Zhiyuan Zhang, Guangxiang Zhao, and Junyang Lin.
\newblock Understanding and improving layer normalization.
\newblock 2019.

\bibitem{r20He2020-hc}
Kaiming He, Haoqi Fan, Yuxin Wu, Saining Xie, and Ross Girshick.
\newblock Momentum contrast for unsupervised visual representation learning.
\newblock In {\em 2020 {IEEE/CVF} Conference on Computer Vision and Pattern
  Recognition ({CVPR})}. IEEE, 2020.

\bibitem{r22Tonekaboni2021-sn}
Sana Tonekaboni, Danny Eytan, and Anna Goldenberg.
\newblock Unsupervised representation learning for time series with temporal
  neighborhood coding.
\newblock 2021.

\bibitem{r23Van_den_Oord2018-ua}
Aaron van~den Oord, Yazhe Li, and Oriol Vinyals.
\newblock Representation learning with contrastive predictive coding.
\newblock 2018.

\bibitem{r24Saxena2008-vr}
Abhinav Saxena, Kai Goebel, Don Simon, and Neil Eklund.
\newblock Damage propagation modeling for aircraft engine run-to-failure
  simulation.
\newblock In {\em 2008 International Conference on Prognostics and Health
  Management}. IEEE, 2008.

\bibitem{r25Jiang2022-ib}
Fei Jiang, Kang Ding, Guolin He, Huibin Lin, Zhuyun Chen, and Weihua Li.
\newblock Dual-attention-based multiscale convolutional neural network with
  stage division for remaining useful life prediction of rolling bearings.
\newblock {\em IEEE Trans. Instrum. Meas.}, 71:1--10, 2022.

\bibitem{LSTM_Base}
Mohamed Sayah, Djillali Guebli, Zeina Al~Masry, and Noureddine Zerhouni.
\newblock Robustness testing framework for rul prediction deep lstm networks.
\newblock {\em ISA transactions.}, 113:28--38, 2021.

\end{thebibliography}

\bio{}
\endbio


\end{document}